\documentclass{article}

\usepackage{arxiv}

\usepackage[utf8]{inputenc} % allow utf-8 input
\usepackage[T1]{fontenc}    % use 8-bit T1 fonts
\usepackage{hyperref}       % hyperlinks
\usepackage{url}            % simple URL typesetting
\usepackage{booktabs}       % professional-quality tables
\usepackage{amsfonts}       % blackboard math symbols
\usepackage{nicefrac}       % compact symbols for 1/2, etc.
\usepackage{microtype}      % microtypography
\usepackage{lipsum}
\usepackage{graphicx}
\graphicspath{ {./figures/} }

\usepackage{hyperref}
\usepackage{tcolorbox}
\usepackage{amsmath}
\usepackage{multirow}
\usepackage{listings}

\makeatletter
\def\xfootnote{\xdef\@thefnmark{}\@footnotetext}
\makeatother

\newcommand{\numberofdatasets}{109 }

\title{Modeling Generalization in Machine Learning:\\A Methodological and Computational Study}
\author{ 
    Pietro~Barbiero\hspace{1mm}\href{https://orcid.org/0000-0003-3155-2564}{\includegraphics[scale=0.06]{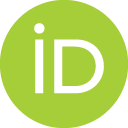}} \\
    Cambridge University\\
    United Kingdom\\
    \texttt{barbiero@tutanota.com} \\
    \And
    Giovanni~Squillero\hspace{1mm}\href{https://orcid.org/0000-0001-5784-6435}{\includegraphics[scale=0.06]{orcid.png}} \\
    Politecnico di Torino\\
    Italy\\
     \texttt{squillero@polito.it} \\
    \And
    Alberto~Tonda\hspace{1mm}\href{https://orcid.org/0000-0001-5895-4809}{\includegraphics[scale=0.06]{orcid.png}} \\
    Universit\'{e} Paris-Saclay, INRAE\\
    France\\
    \texttt{alberto.tonda@inrae.fr}
}

\begin{document}
\maketitle
\xfootnote{Authors are listed in alphabetical order}

\begin{abstract}
As machine learning becomes more and more available to the general public, theoretical questions are turning into pressing practical issues. Possibly, one of the most relevant concerns is the assessment of our confidence in trusting machine learning predictions. In many real-world cases, it is of utmost importance to estimate the capabilities of a machine learning algorithm to generalize, i.e., to provide accurate predictions on unseen data, depending on the characteristics of the target problem. %In many real-world cases, machine learning may experience difficulties unless it cannot be explained how the model is working and under which assumptions it will be able to generalize, i.e., to provide accurate predictions on unseen data. %
In this work we perform a meta-analysis of \numberofdatasets publicly-available classification data sets, modeling machine learning generalization as a function of a variety of data set characteristics, ranging from number of samples to intrinsic dimensionality, from class-wise feature skewness to $F1$ evaluated on test samples falling outside the convex hull of the training set. Experimental results demonstrate the relevance of using the concept of the convex hull of the training data in assessing machine learning generalization, by emphasizing the difference between interpolated and extrapolated predictions. Besides several predictable correlations, we observe unexpectedly weak associations between the generalization ability of machine learning models and all metrics related to dimensionality, thus challenging the common assumption that the \textit{curse of dimensionality} might impair generalization in machine learning.
\end{abstract}

% keywords can be removed

\section{Introduction}
\label{sec:introduction}
\label{S:1}

The term \emph{machine learning} (ML) traditionally includes algorithms that are able to improve their performance on a specific task over time, given an increasing amount of relevant data~\cite{Mitchell1997Machine}. In recent years, this field of research is enjoying a growing popularity, driven by the breakthrough of Deep Learning~\cite{lecun2015deep} and an impressive track record of success stories in different fields, ranging from natural language processing \cite{mikolov2013distributed} to autonomous vehicles~\cite{chen2015deepdriving}, image classification~\cite{krizhevsky2012imagenet} human-competitive performance in boardgames~\cite{silver2017mastering}. An interesting \emph{online collection} about various uses of ML has been compiled by the journal Nature is late 2018\footnote{https://www.nature.com/collections/csgqqsrfxh/content/machine-learning (\emph{The multidisciplinary nature of machine intelligence}, 26 September 2018)}, although the fast pace the field is progressing made it to appear outdated after few quarters.

As out-of-the-box ML solutions are becoming increasingly available to both researchers and the general public \cite{pedregosa2011scikit,eibe2016weka,Datarobot,chollet2015keras}, theoretical questions are suddenly turning into practical issues. Among all common inquiries, perhaps the most basic is: can ML work on a specific problem? Or, in other words: given the characteristics of a target data set, can the effectiveness of a ML approach be predicted? Interestingly, this latter question can be further rephrased as: what are the characteristics of a data set that are well correlated with the possibility, or the impossibility, of obtaining ML models able to effectively extrapolate to unknown instances of the problem? It is well known that ML algorithms are affected by the \emph{curse of dimensionality} \cite{altman2018curse}, but ML practitioners also know that it could be possible to obtain reliable models even for high-dimensional data sets, and with a relatively small number of samples \cite{barbiero2018understanding}. The common approach among practitioners in the field, when dealing with a new data set, seems to be: try as many different ML algorithms as possible in a cross-validation, and evaluate the outcomes; then focus on the techniques that provided the best results, possibly applying them in an \emph{ensemble} \cite{altman2017points}.

Taking inspiration from \cite{Oreski2017}, where the authors find links between data set characteristics and efficiency of feature selection techniques, we propose to empirically explore the relation between data-set characteristics and effectiveness of standard ML models, in order to obtain a general meta-model able to extrapolate. In order to answer the question, we analyzed \numberofdatasets publicly available classification data sets from open-access, curated sources. We decided to focus on classification, as supervised ML represents a quite significant portion of real-world problems; and, differently from regression, several sophisticated quality metrics have already been developed for this task~\cite{Michie:1995:MLN:212782}.

During the analysis, we take into account characteristics such as number of features, number of classes, number of samples, and we look for correlations with quality metrics, such as accuracy of a ML model on training and test points. Extrapolation is assessed not just by alternatively dividing the data into training and test sets, but by analyzing whether data points fall inside or outside of the convex hull of the training data. After collecting the meta-data on the performance of a state-of-the-art classification algorithm on the data sets, the statistical analysis presents both predictable and surprising results, hinting at the fact that dimensionality might not be so cursed after all. 

%Let's take this paper~\cite{Oreski2017} and try to get some inspiration.

\bigskip
\begin{tcolorbox}[colback=blue!5!white,colframe=blue!75!black,title=Main contributions]
\begin{enumerate}
    \item We ran a quantitative evaluations of ML models over \numberofdatasets publicly available data sets.
    \item In Section \ref{sec:course:the_curses_of_dimensionality} and Section \ref{sec:extrapolation} we provide a general overview on generalization in ML, and of the so-called \emph{curse of dimensionality}.
    \item In Section \ref{sec:chull} we propose to take into consideration the convex hull of the training set to better understand generalization in ML, as it makes it possible to separate cases when an algorithm interpolates or extrapolates on test points.
    \item In Section \ref{sec:methodology} we empirically model associations between the ability to generalize and the data set characteristics, challenging the relationship between generalization ability and the dimensionality.
\end{enumerate}
\end{tcolorbox}
\bigskip

\section{The curses of dimensionality}
\label{sec:course:the_curses_of_dimensionality_and_extrapolation}
\label{sec:course:the_curses_of_dimensionality}
\label{S:2}

%In this section, we present a summary of the concept of \emph{extrapolation} in ML, we briefly discuss possible options to evaluate extrapolation capabilities of a ML model, and we discuss the curse of dimensionality.\todo{conventional wisdom: dimensionality is cursed}
%    \item New papers: \sout{Doshi-Velez and Kim - 2017 - Towards A Rigorous Science of Interpretable Machin}, \sout{Fawzi et al. - 2017 - Classification regions of deep neural networks}, Geirhos et al. - Generalisation in humans and deep neural networks, Li et al. - 2018 - On the Decision Boundary of Deep Neural Networks, Morcos et al. - 2018 - On the importance of single directions for general, Neyshabur et al. - 2018 - Towards Understanding the Role of Over-Parametriza, Ramamurthy et al. - 2018 - Topological Data Analysis of Decision Boundaries w, Wu et al. - 2017 - Towards Understanding Generalization of Deep Learn, \sout{Zhang et al. - 2017 - Understanding Deep Learning Requires Re-Thinking Generalization}

% \subsection{The curse of dimensionality}

%The concept of \emph{extrapolation} in ML is indissolubly linked with the  

The \emph{curse of dimensionality} denotes a variety of different phenomena that impair data analysis if a large number of variables need to be considered at the same time \cite{bellman1966dynamic, bellman2015adaptive}. While in most cases it has no closed form nor a unique solution, it has distinctive deleterious effects. Problems like \emph{data sparsity}, \emph{collinearity}, and \emph{overfitting}, seem to confirm the platitude that in ML dimensionality is cursed~\cite{altman2018curse}. 

\begin{figure}[htb]
    \centering
    \includegraphics[width=0.49\textwidth]{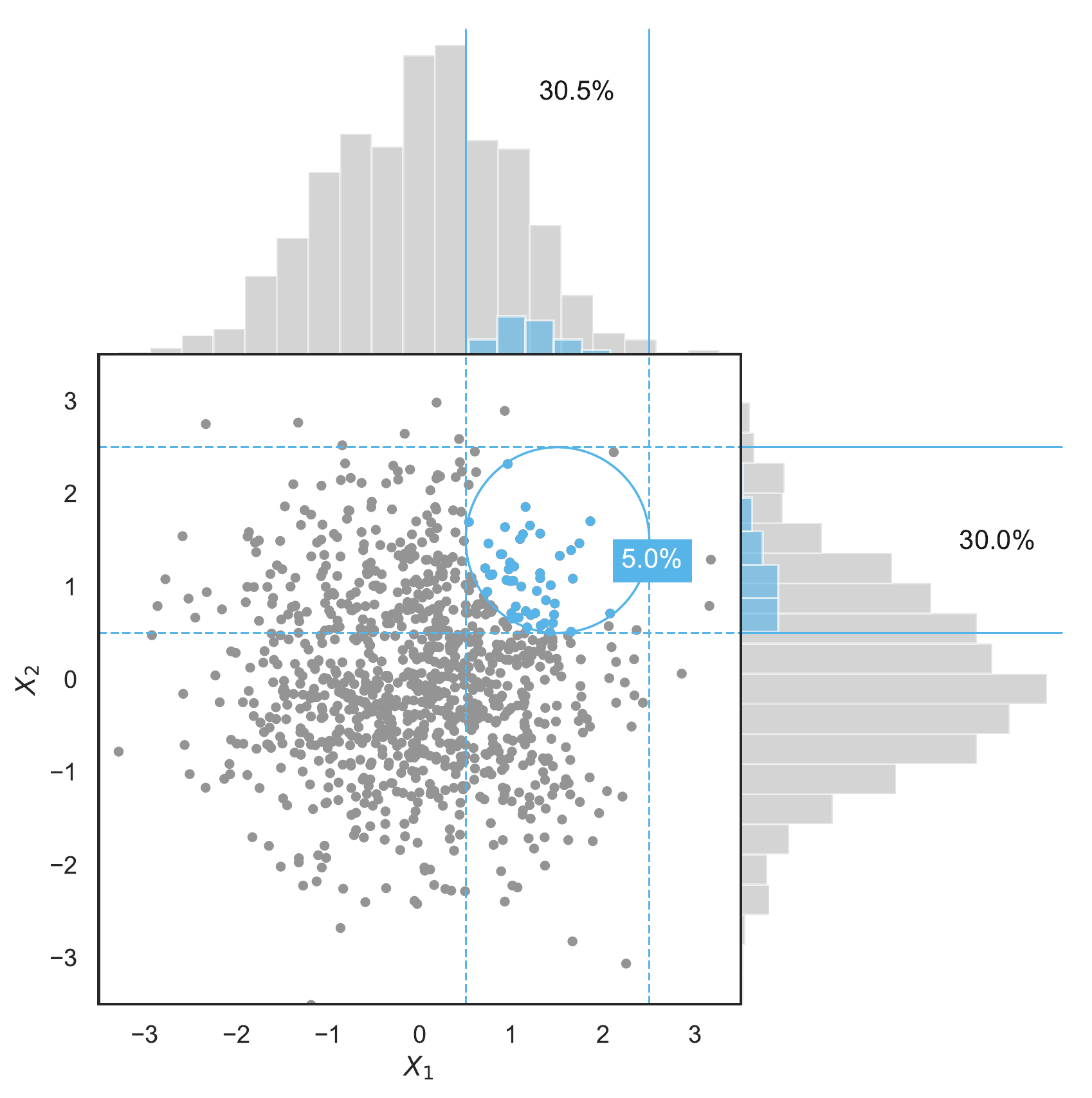}
    \includegraphics[width=0.49\textwidth]{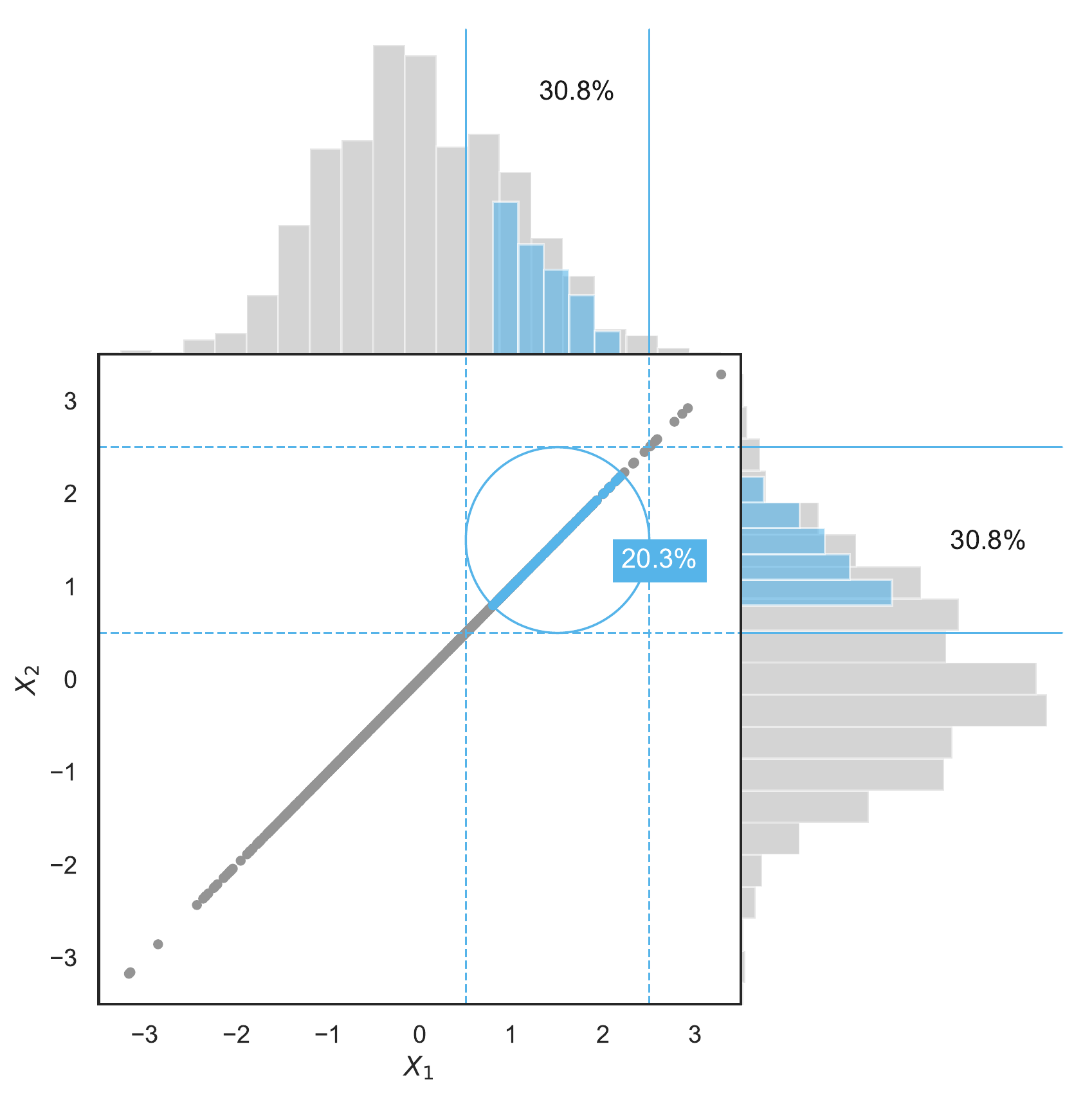}
    \caption{Visual representation of the data sparsity phenomenon in case of correlated random variables (left) and collinear random variables (right).}
    \label{fig:data-sparsity}
\end{figure}

As an example, let consider a collection of $n$ data points generated by sampling two random variables $X_1$ and $X_2$ originated from two standard normal distributions ($\mu=0, \sigma=1.0$). The amount of samples falling within the interval $x_i \pm \sigma$ with $x_i = 1.5$ is around the $30\%$ both for $X_1$ and $X_2$, individually (grey histograms in Figure \ref{fig:data-sparsity}). However, when the joint distribution of $X_1$ and $X_2$ is considered, the amount of samples falling within the joint interval $(x_1 \pm \sigma, x_2 \pm \sigma)$ with $x_{1,2} = 1.5$ drops. The worst-case scenario arises when the two random variables are uncorrelated (Figure \ref{fig:data-sparsity}, left). As the number of random variables ($p$) increases, the fraction of points within the $p$-dimensional interval of radius $\sigma$ decreases rapidly: $\sim 5\%$ for $p=2$, $\sim 1\%$ for $p=3$, and $\sim 0\%$ for $p = 4$ \cite{altman2018curse}. In practice, the sparser the samples the harder will be collecting data that are representative of the population. The best-case scenario occurs when the random variables are perfectly correlated (Figure \ref{fig:data-sparsity}, right). In this case, the decrease is still significant, but much slower (from $30\%$ to $20\%$ rather than $5\%$ for $p=2$). 

In most real-world scenarios, a considerable number of variables are correlated. As supervised ML algorithms are usually devised to select and use variables strongly correlated with the target variable, the data-sparsity problem may be usually mitigated by considering together highly correlated sets of variables. However, exploiting correlated variables may also have a catastrophic impact when variables are used for prediction: as there could be more than one subset of variables yielding approximately the same result, considering them together can make to make impossible to understand the individual impact of each variable. For example, suppose that two variables $X_1$ and $X_2$ can be used to predict a target variable $Y$ by means of a predictive model $f$:

\begin{equation}
Y = f(X_1, X_2)
\end{equation}

Now suppose that there exists another variable $X_3$ that can be expressed as a function of both $X_1$ and $X_2$, e.g., $X_3 = X_1 + 2 X_2$. Such a system of equations can perfectly be solved by using an arbitrary pair of variables $\{(X_1, X_2), (X_1, X_3), (X_2, X_3)\}$, as they are all perfectly correlated. Even though the predictive problem appears to be solved, the causative source of variability of $Y$ is now uncertain, and the relative importance of the variables cannot be estimated from data. The situation gets critical when the number of variables exceeds the number of samples ($p > n$): at least one of the variables can always be expressed as a linear combination of the others, thus yielding multiple perfect correlations \cite{zimek2012survey}.

In case of multicollinearity, one of the possible solutions exploited by supervised ML algorithms (e.g., logistic regression \cite{mckelvey1975statistical}) is to associate a weight to each variable, corresponding to its relevance. Instead of discarding variables which may be the true causative source of variability, these approaches make it possible to take into account all the observed variables at the same time, by increasing the number of model's parameters. Such increase in model complexity may in turn be a possible cause of overfitting, another phenomenon related to the curse of dimensionality. In fact, the increased flexibility makes the model not only able to fit the underlying relationship between variables, but also the random idiosyncrasies of the observations \cite{lever2016pointsmodel}.

\begin{figure}[!ht]
    \centering
    \includegraphics[width=0.49\textwidth]{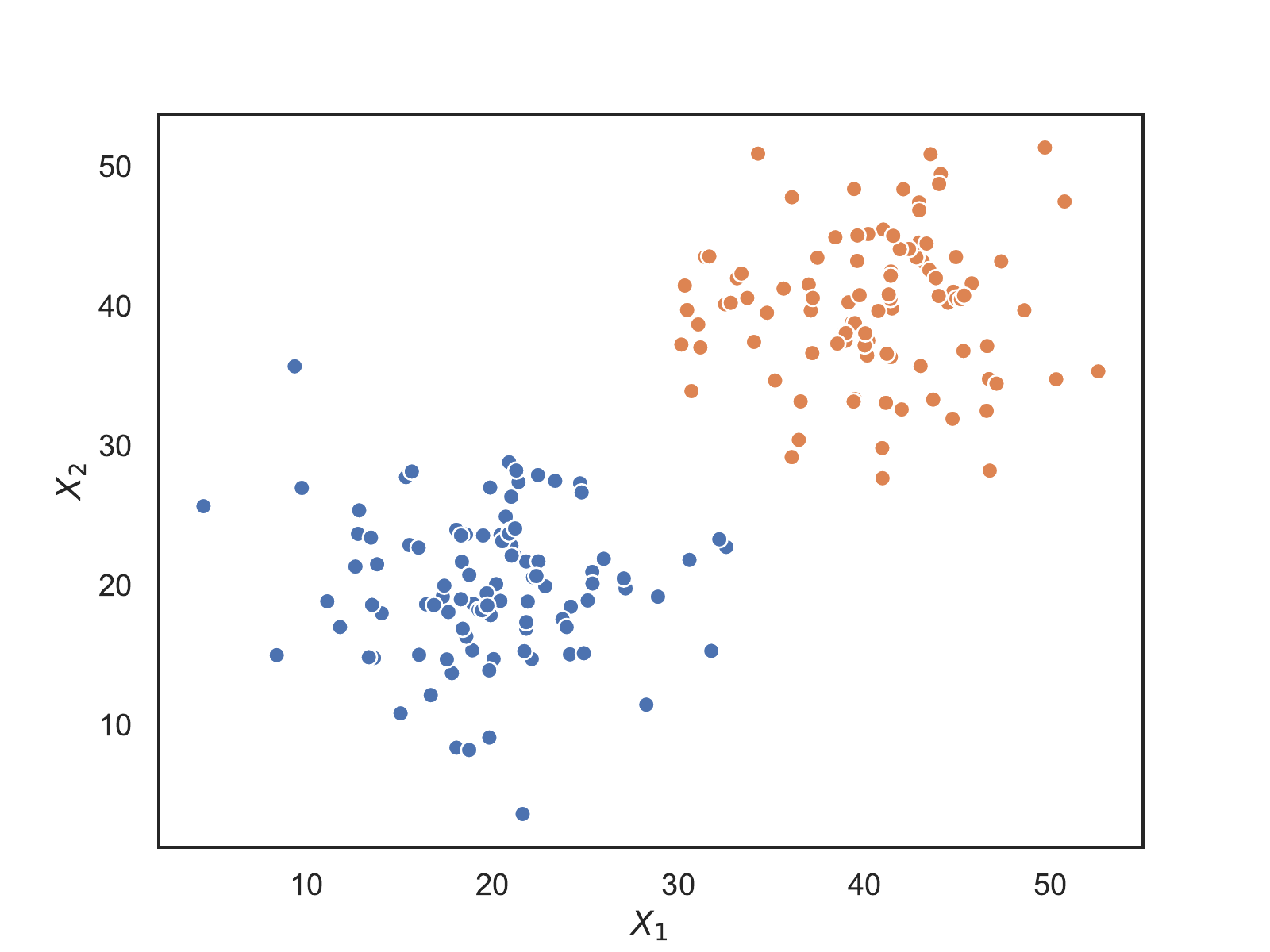}
    \includegraphics[width=0.49\textwidth]{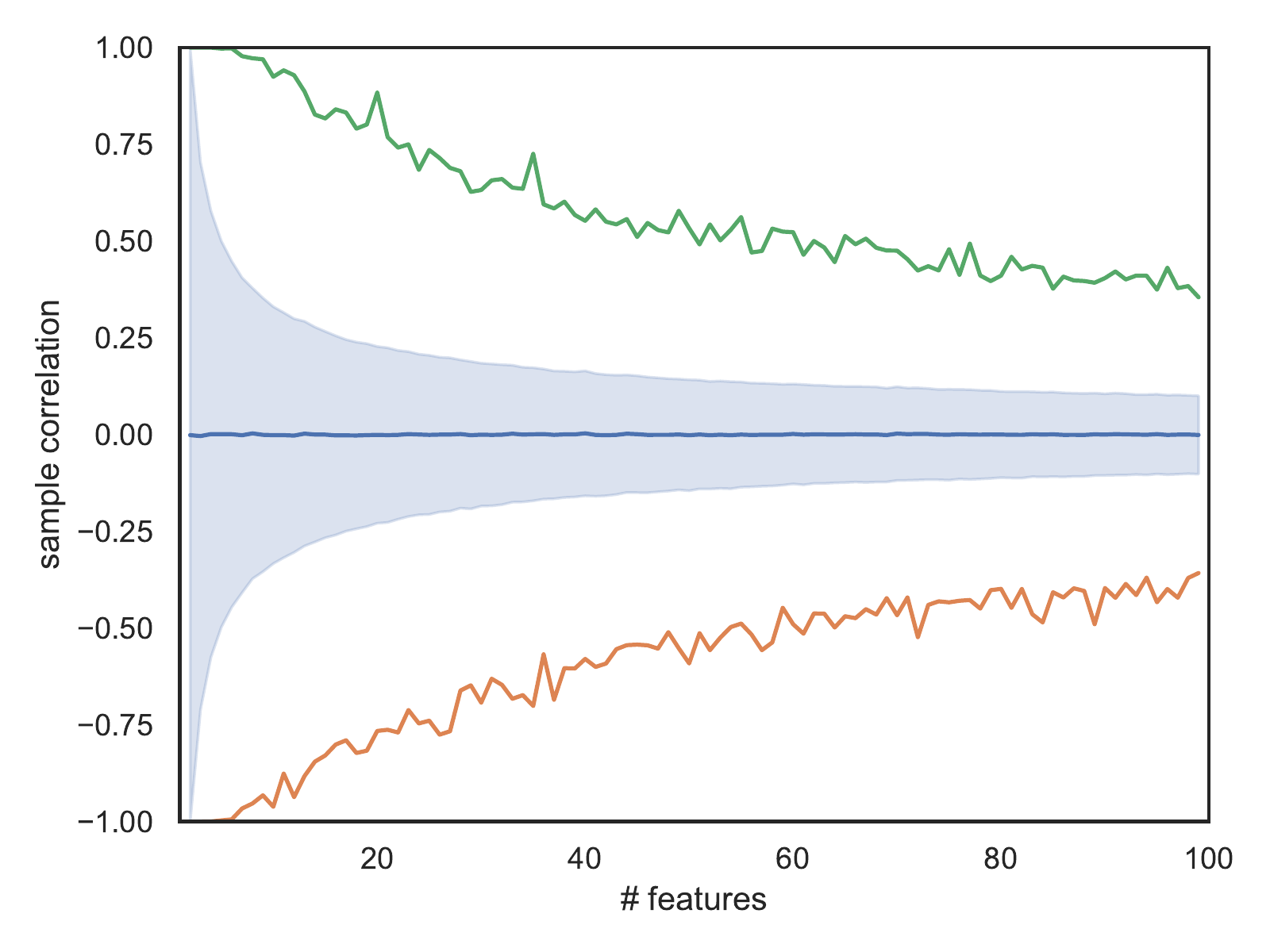}\\
    \includegraphics[width=0.49\textwidth]{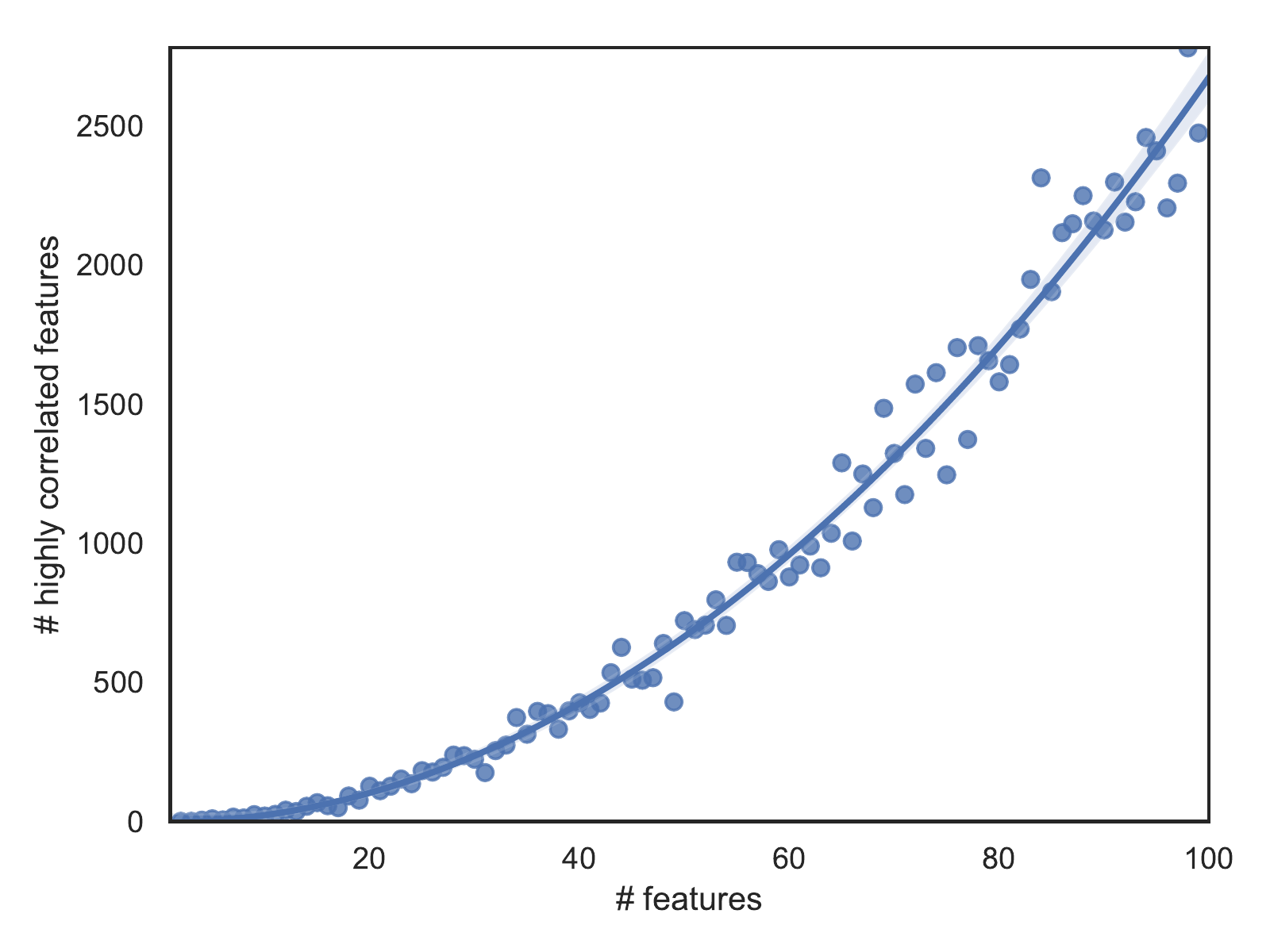}
    \includegraphics[width=0.49\textwidth]{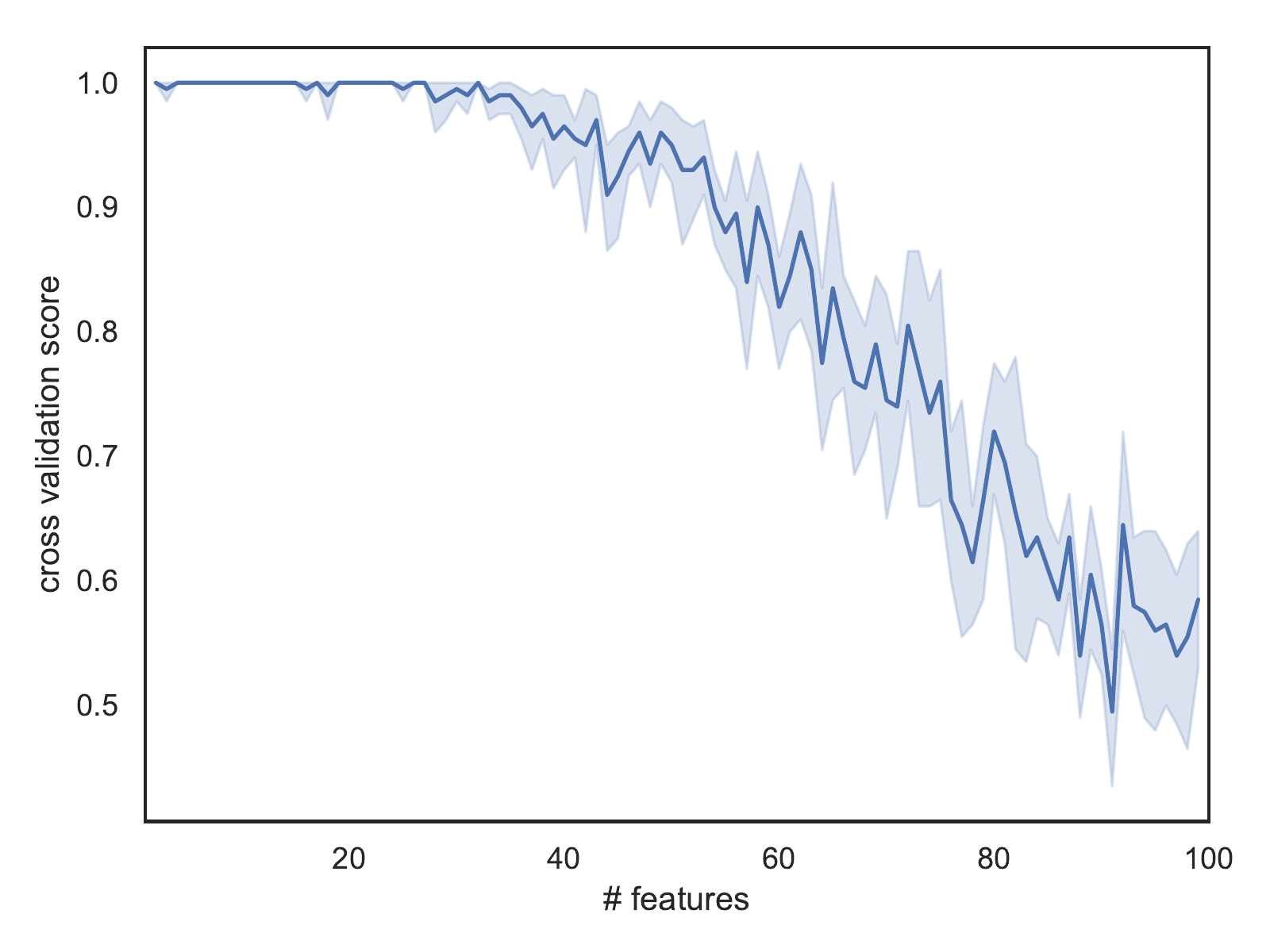}
    \caption{Two-class samples drawn from two correlated Gaussian random variables. Notably the classification problem is trivial as the two classes are linearly separable (top-left). By increasing the number of correlated random variables the highest value (in green), the lowest value (in orange), and the variance of samples' correlation shrink towards zero (top-right). The number of highly correlated variables ($\rho > 0.8$) increases polynomially with the overall number of features (bottom-left). On the other hand, the generalization accuracy of logistic regression in predicting class labels decreases, down to the point of providing almost random estimates (bottom-right).}
    \label{fig:cods-issues}
\end{figure}

\begin{figure}[htb]
    \centering
    \includegraphics[width=0.8\textwidth]{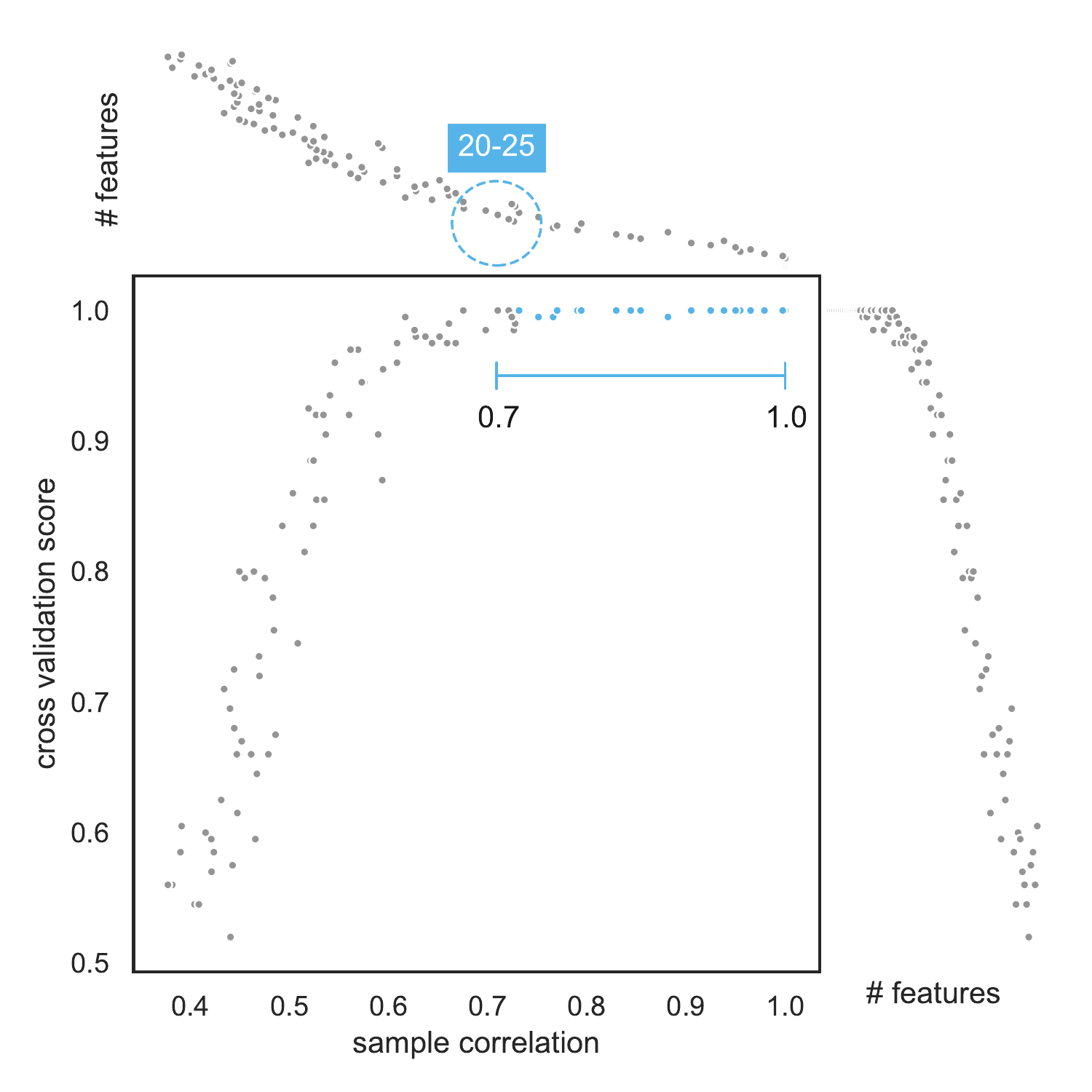}
    \caption{Mutual relationships among generalization ability (cross-validation accuracy), sample correlation, and number of features when all variables are correlated and gaussian. When the number of features is higher than 20--25, samples' correlation drops leading to a dramatic decrease in generalization ability.}
    \label{fig:cods-relations}
\end{figure}

\section{Extrapolation and interpolation in machine learning}
\label{sec:extrapolation}

The objective of supervised ML can be roughly summarized as automatically obtaining a \emph{predictive model} for a specific phenomenon starting from a set of known cases. Usually such known cases consist in different instances of measurements, or \emph{samples}, each one specifying the values of all different variables, or \emph{features} of the problem. One of these features is the \emph{target variable}, that is, what should be predicted by the final model. ML algorithms try to find a mathematical relationship between the target feature and the others in the set of known data, or \emph{training set} of data. Eventually, such relation can be encoded with a linear model~\cite{lever2016pointslogistic}, more complex structures~\cite{lecun2015deep}, or even ensembles of simpler models~\cite{altman2017points,krzywinski2017points}. Indeed, it is quite important to verify to what extent the model obtained is able to \emph{generalize}, that is, to provide meaningful prediction for new instances of the problem. A model that is only able to deal with the data it was trained on is said to be \emph{overfitted}~\cite{lever2016pointsmodel}, and it is generally considered useless regardless its performance on the train set.

%The concept of overfitting is strictly linked to the capacity of a given model.

% I don't know where the following paragraph can be placed
In ML the term \emph{capacity} may describe the ability of a model to represent complex relationships --- the term \emph{complexity}, when referred to a model, can also be used with a similar meaning. While it sounds obvious that a second-degree polynomial has a higher capacity than a linear regression, and can better fit more instances of data, there are relatively few contributions in literature that attempt to provide a more formal framing~\cite{vapnik2015uniform,bartlett2002rademacher,poggio2004general}, and these terms are often used in rather intuitive and qualitative statements. As a fast but gross approximation, ML practitioners often assess model capacity by evaluating the number of parameters that can be tuned inside a model. In theory, the best ML model for a task is the one with just enough capacity to properly represent the training data: Models with lower capacity would \emph{underfit}, i.e., they deliver unsatisfying results as they are unable to cope with the complexity of the phenomenon; models with higher capacity would risk to \emph{overfit} and consequently generalize poorly. 

\begin{figure}[htb]
    \centering
    \includegraphics[width=0.99\textwidth, trim=4.2cm 0cm 3.5cm 0cm, clip]{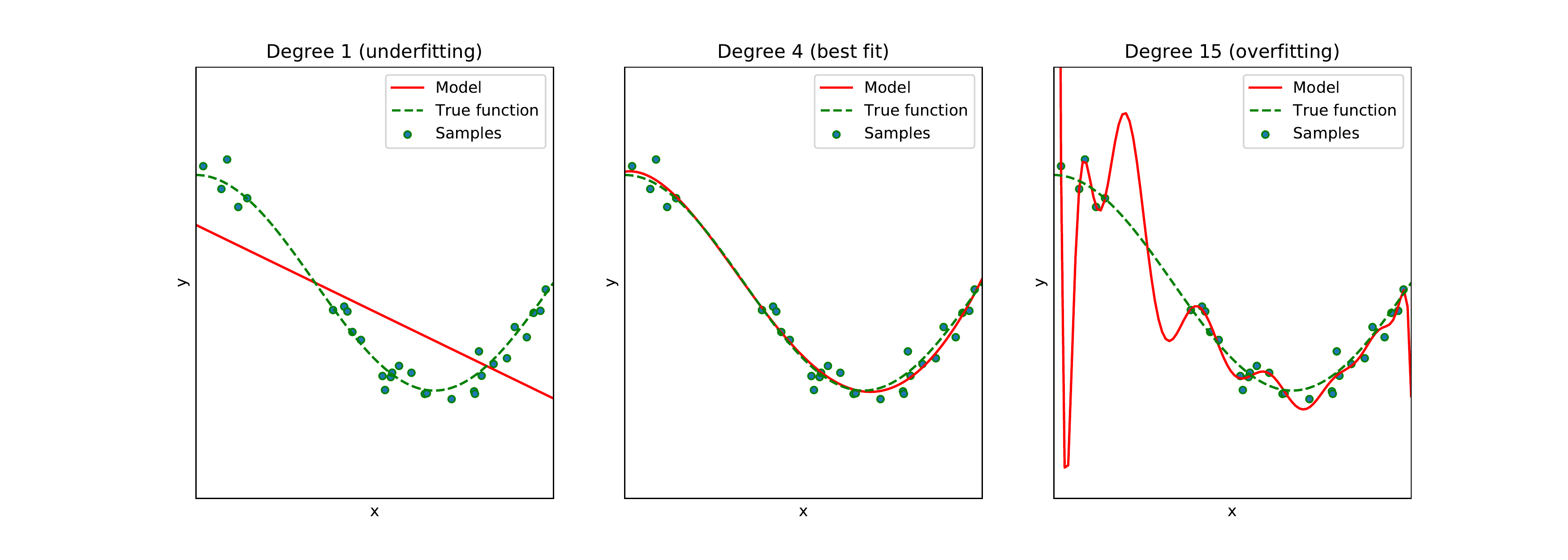}
    \caption{Visual representation of the effect of model capacity on fitting. The original data points can be properly represented by a polynomial of degree 4 (\textbf{middle}), so a polynomial regression with lower capacity will underfit (\textbf{left}), while a polynomial regression with higher capacity will overfit (\textbf{right}).}
    \label{fig:overfitting-underfitting}
\end{figure}

A simple depiction of overfitting and underfitting is provided in Figure~\ref{fig:overfitting-underfitting}. In practice, unfortunately, it's extremely hard to estimate the capacity necessary to represent a data set; and the solution that many ML practitioners use is --- yet again --- to apply several techniques with increasing capacity to the data set, until either the gain in fitting stops, or the improvements are not considered important enough to justify an increase in capacity. More interestingly, even estimating effective model capacity is not trivial, as there is evidence from works on deep learning that models with enough parameters to theoretically overfit the training data are actually able to generalize well in real-world case studies \cite{zhang2016understanding}. The trade-off between fitting and capacity has been independently explored by different ML communities, with the definition of model-dependent metrics that attempt to take into account both fitting and capacity to assess overall quality, to facilitate model selection: A few examples include the Akaike Information Criterion \cite{akaike1974new}, the Bayesian Information Criterion \cite{schwarz1978estimating}, and Pareto-based approaches used mainly in symbolic regression \cite{smits2005pareto}. 

While evaluating model capacity can help reducing the chance of overfitting,  \emph{measuring} overfitting remains far from trivial. Ideal ML models should be able to obtain good predictions even for \emph{unknown} samples of the same problems, but - by definition - the models cannot be tested on unknown data sets. Given this practical need, ML researchers found ways to at least \emph{assess} overfitting, through different techniques. A basic, but extremely efficient technique, is cross-validation (with all its variations, such as leave-one-out cross-validation, stratified cross-validation and the like): The training data is split into $k$ folds of equal size, a ML algorithm is iteratively trained on all folds minus one, and tested on the remaining fold \cite{stone1974cross}. Analyzing the results of a $k$-fold cross-validation, for example the average performance, or single instances where the performance on a test fold differs greatly from the others, can provide further insight on the problem characteristics. 

\begin{figure}[htb]
    \centering
    \includegraphics[width=0.99\textwidth, trim=2.2cm 0cm 2cm 0cm, clip]{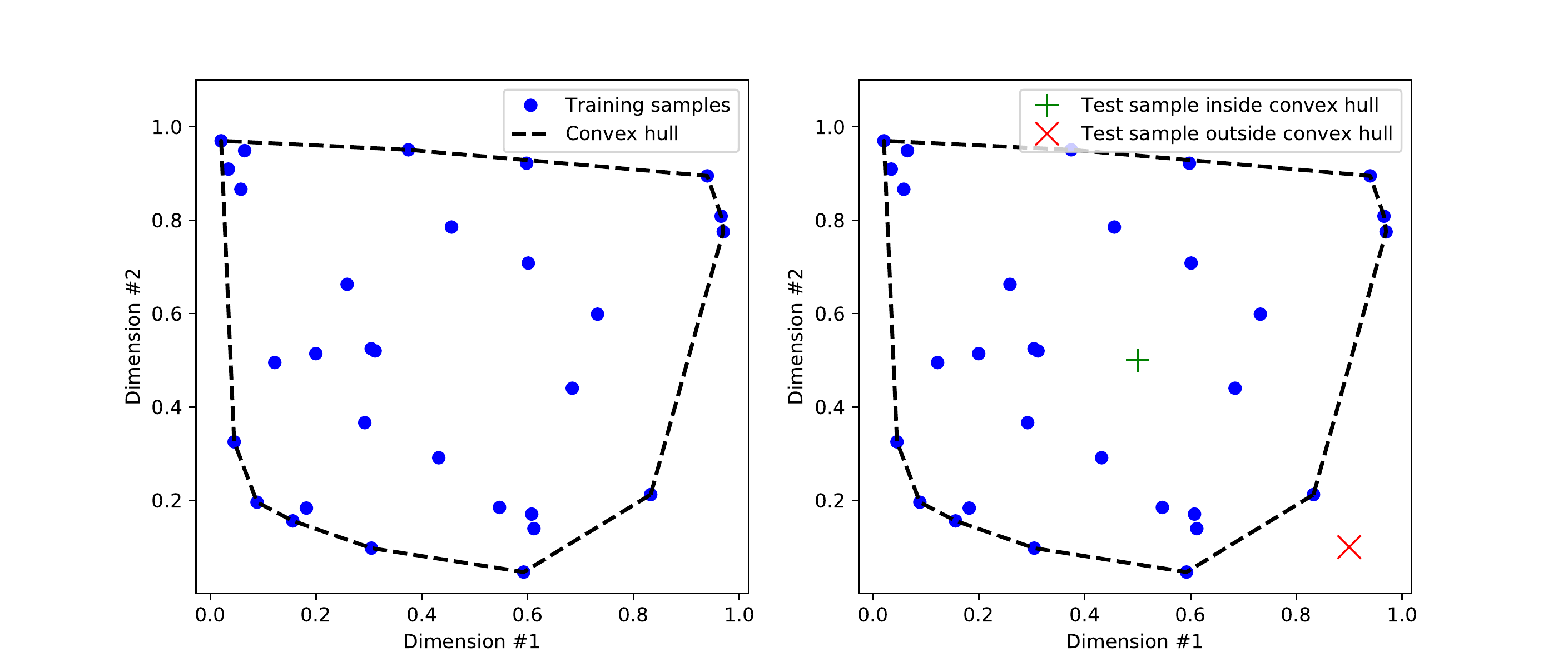}
    \caption{An example of convex hull. The convex hull of a set of training samples is the minimal hyper-polygon (in this case, a 2D polygon) that contains all the data inside its edges (left). For a machine learning model trained on the initial training set, predicting a value for an unseen test point inside the convex hull (in green) probably requires interpolation; but predicting for a test point outside the convex hull requires extrapolation (right).}
    \label{fig:convex-hull-example}
\end{figure}

As the real objective of evaluating overfitting is to assess a model's capability of extrapolating to unknown instances of the same problem, it is worth it to spend a few words on the meaning of \emph{extrapolation} in supervised ML. As for model capacity, there is an intuitive and imprecise concept of extrapolation, defined as the ability of the model to correctly predict data points that are \emph{considerably different} from the information provided in the training data, but still belong to the same problem. An alternative outlook on extrapolation comes from computational geometry: Interpreting the data points in the training set as points in $\mathbb{R}^f$, where $f$ is the number of features, it is possible to compute its \emph{convex hull}, the smallest polygon that contains all training points. Given the convex hull of the training set, it is then possible to assess whether an unseen test data point will fall inside or outside the convex hull. It is reasonable to assume that, for points inside the convex hull, a ML model will \emph{interpolate} known data to obtain a prediction; while the same model will \emph{extrapolate} for test points placed outside of the convex hull. An example is presented in Figure \ref{fig:convex-hull-example}. It is important to notice that, depending on the characteristics and the distribution of the training points, this interpretation of interpolation/extrapolation might not correspond to the actual difficulty of predictions for the model. For example, it is possible to imagine a situation where the model will provide better predictions for of a test point outside of the convex hull of the training data, but still relatively close to known points, than for a test point located inside the convex hull of the training data, but in a part of the space where training points are relatively sparse. Still, in most practical scenarios, it is generally harder for models to reliably predict values for test points outside of the convex hull of the training data. A more in-depth discussion on the convex hull is provided in the following Section.

\section{Assessing generalization tasks using the convex hull}
\label{sec:chull}

In an Euclidean space, the convex hull of a set of points $X = \{x_i \in\mathbb{R}^{d}\}$ is the smallest convex set containing all the points in $X$. If the number of points $n$ in $X$ is finite (i.e. if $X$ is a matrix $\mathbb{R}^{n \times d}$), then the convex hull forms a convex polytope in $\mathbb{R}^d$. Finding faces or the set of extreme points of such convex polytope is a NP-hard problem \cite{tiwary2008hardness}. However, checking if a point $z$ lies inside or outside the polytope is much easier, and can be performed in polynomial time~\cite{spielman2004smoothed}. When dealing with ML models, evaluating the convex hull of a training set can provide extra information on unseen test points: If a test point falls inside the convex hull, it is expected that the ML model will probably \emph{interpolate} known points to find the predicted value; vice versa, if a test point falls outside the convex hull of the training set, the ML model will likely be required to \emph{extrapolate} to obtain a prediction. 

The problem of checking whether a point $z$ lies inside the polytope of $X$ has a simple linear programming formulation \cite{pardalos1995linear}:

\begin{equation}
    \begin{aligned}
    & \underset{y}{\text{min}}
    & & c^T y \\
    & \text{s.t.} & &  Ay = b \\
    & & &  c,y \in \mathbb{R}^n. \\
    \end{aligned}
\end{equation}

where:
\begin{equation}
    c = \boldsymbol{0}
    \qquad
    A = 
    \begin{bmatrix}
        X^T\\
        \boldsymbol{1}^T
    \end{bmatrix}
    \qquad
    b = 
    \begin{bmatrix}
        z\\
        1
    \end{bmatrix}
\end{equation}

Such formulation is  known as a \emph{Phase I} method \cite{boyd2004convex}, as the final goal is not the actual optimization of variable $z$, but rather checking whether a feasible solution does exist. In such contexts, the cost function can be a constant, as the only objective is satisfying the constraints. The first $n$ constraints impose that the position of $z$ in the feature space must be a combination of the points $X$:
\begin{equation}
    z = \sum_{i=1}^n y_i x_i
%\label{eq:z}
\end{equation}
The last constraint imposes that such combination must be convex, which implies, by definition, that the coefficients $y_i$ must sum to $1$:
\begin{equation}
    \sum_{i=1}^n y_i = 1
\end{equation}
If the Phase I problem is feasible, point $x$ can be expressed as a convex combination of the set of points $X$. By definition, this means that point $z$ lies inside the convex polytope of $X$. On the contrary, if the Phase I problem does not have any feasible solution, then point $z$ lies outside the convex hull of $X$.

Even the proposed approach exploiting the convex hull can be affected by the curse of dimensionality. The critical point occurs when the dimension of the Euclidean space exceeds the number of observations ($d > n$). In this case the upper bound of the rank of the matrix $X$ corresponds to the number of samples $n$ \cite{mirsky2012introduction, mackiw1995note}:

\begin{equation}
rank(X) \leq n
\end{equation}

Therefore, the maximal number of linearly independent columns of $X$ cannot be higher than $n$. Whatever the number of dimensions $d$, the points $x_i$ will lie in a subspace $\mathbb{R}^s$ where $s \leq rank(X) \leq n < d$. As a consequence, the convex hull generated by the set of points $x_i$ will belong to the same subspace. By definition, for $s < d$, the subspace $\mathbb{R}^s$ has \textit{measure zero} in $\mathbb{R}^d$ and can be considered as negligible \cite{folland2013real}. Hence, when a new point $z$ is added to the space, it is \textit{almost sure} that it will fall outside negligible subspaces as $\mathbb{R}^s$ \cite{jacod2012probability}. In summary, when $d > n$, the new point $z$ will \textit{almost never} belong to the convex hull of $X$, making the computation of the convex hull ultimately useless.

\section{Modeling generalization as a function of data set characteristics}
\label{sec:methodology}

The objective of this work consists in assessing generalization abilities of machine learning models with empirical experiments. To this aim, first we selected (i) a large set of publicly available data sets for classification, (ii) a representative set of machine learning classifiers, and (iii) a set of data set characteristics. Then, on each data set we perform a cross-validation using ML models and we compute relevant metrics for each fold. Finally, we analyze the relationships between data set characteristics and generalization ability of ML models using both classical correlation metrics, and association models derived through symbolic regression.

In the following subsections, the methodology adopted for selecting the data sets and the corresponding metrics to characterize them are introduced and discussed.

%\begin{itemize}
%\item How did we pick the data sets? Some of them just do not work. Out of potentially 2,500 data sets we selected just classification problems, and discarded ones that were not accessible.
%\item Pariamoci il culo con i bias di cui sotto, vedi Conclusions
%\end{itemize}

\subsection{Data sets}
Following the analyses presented in~\cite{Oreski2017} and~\cite{Michie:1995:MLN:212782}, this study focuses on classification, as it is easier to characterize classification rather than regression data sets, as several of the characteristics analyzed are based on comparing batches of samples belonging to different classes (see \emph{Statistical measures} paragraph in subsection \ref{Ss:data set}). Additionally, as the following analysis is partly based on evaluating convex hulls, only data sets with real-valued features are considered. 

The data sets examined are acquired from the OpenML repository~\cite{OpenML2013}, on online, curated collection that, as the time of writing, includes over 3,100 data sets of different kinds. After selecting only data sets related to classification problems, with real-valued features exclusively, and discarding those  containing errors, a total of \numberofdatasets data sets was ultimately retained for the analysis. The number of samples in the selected collection ranges from $47$ to $44,819$, while the number of features spans from $2$ to $3,758$. All selected data sets have real-valued features and a discrete target (suited for classification). The mean feature correlation of the data sets is $0.619$, with a standard error of the mean of $0.007$, the average intrinsic dimensionality ratio is $0.629$, with a standard error of the mean of $0.047$.

\subsection{Data set characterization}
\label{Ss:data set}
As in previous meta-analyses \cite{Oreski2017,Michie:1995:MLN:212782}, each data set is characterized using metrics, grouped into four categories: simple, statistical, Euclidean, and generalization metrics. For each data set, such measures are computed over a stratified 10-fold cross validation~\cite{stone1974cross}.

% Please add the following required packages to your document preamble:
% \usepackage{multirow}
\begin{table}[!ht]
\renewcommand{\arraystretch}{1.2}
\centering
\caption{Summary of the metrics used to characterize the data sets analyzed in the study.}
\label{tab:my-table}
\resizebox{\textwidth}{!}{%
\begin{tabular}{lll}
\hline
Metric type & Symbol & Description \\
\hline
\multirow{3}{*}{\textit{Standard metrics}} & $n$ & Number of samples \\
 & $d$ & Number of features \\
 & $c$ & Number of classes \\
 & & \\
\multirow{5}{*}{\textit{Euclidean metrics}} & $\mathfrak{I}$ & Intrinsic dimensionality \\
 & $\mathfrak{I}_r$ & Intrinsic dimensionality ratio \\
 & $\mathfrak{N}$ & Feature noise ($1-\mathfrak{I}_r$) \\
 & $\mu_{\mathfrak{D}}$ & Average sample distance \\
 & $\sigma_{\mathfrak{D}}$ & Standard deviation of sample distance \\
 & & \\
\multirow{5}{*}{\textit{Statistical metrics}} & $\lambda$ & Average of Levene's test p-values \\
 & $\rho$ & Average of class-wise feature correlation \\
 & $\gamma$ & Average of class-wise feature skewness \\
 & $\kappa$ & Average of class-wise feature kurtosis \\
 & $\eta$ & Average of feature-target mutual information \\
 & & \\
\multirow{9}{*}{\textit{Generalization metrics}} & $CI_{train}$ & Class-imbalance of training samples \\
 & $CI_{test}$ & Class-imbalance of test samples \\
 & $T_{in}$ & Ratio of test samples inside the convex hull \\
 & $T_{out}$ & Ratio of test samples outside the convex hull ($1-T_{in}$) \\
 & $CI_{in}$ & Class-imbalance of test samples inside the convex hull \\
 & $CI_{out}$ & Class-imbalance of test samples outside the convex hull \\
 & $F1_{train}$ & $F1$ for the training set \\
 & $F1_{test}$ & $F1$ for the whole test set \\
 & $F1_{in}$ & $F1$ for the part of the test set inside the convex hull (interpolation ability)\\
 & $F1_{out}$ & $F1$ for the part of test set outside the convex hull (extrapolation ability)\\
\hline
\end{tabular}
}
\end{table}

\paragraph{Simple metrics} Simple metrics describe general characteristics of data sets, namely number of samples ($n$), number of features ($d$, a.k.a. \textit{dimensionality}), and number of classes ($c$)~\cite{daskalaki2006evaluation}.

\paragraph{Statistical metrics} Statistical metrics assess (i) class differences in feature distributions and shapes, and (ii) relationships between features and classification target.

\emph{Levene's test}~\cite{levene1960robust} is an inferential statistical test used to assess if a vector of random variables is \textit{homoscedastic}, i.e. if the variance of the random variables is \textit{almost equal}. In the following, Levene's test is used to compare class covariances for each data set feature. The lower the $p$-value, the higher will be the probability that the class covariances of the feature under study are different \cite{amrhein2017earth}. In the following experiments, the score $\lambda$ collected for each data set is the average of the Levene's $p$-values of its features. 

\emph{Pearson's correlation coefficient}~\cite{pearson1920notes} measures the linear relationship between two variables, providing an indication of the interdependence between pairs of features. The correlations between all pairs of attributes are calculated for each class separately. Since the objective is to evaluate the strength of the relationship and not its sign (positive or negative), the absolute value of the coefficient is used. For each data set, the collected score $\rho$ is the average of the coefficient over all pairs of features and over all classes.

\emph{Skewness}~\cite{pearson1905skew} corresponds to the third standardized moment of a random variable. It indicates the magnitude of the asymmetry of a feature around its mean, yielding an estimate of the feature's departure from normality. The skewness for a class is computed as a weighted average of the skewness of the feature values of its samples. The final skewness score $\gamma$ represents the average skewness over all classes.

\emph{Kurtosis}~\cite{pearson1905fehlergesetz} corresponds to the fourth standardized moment of a random variable. It indicates the "thickness" of the tails of a density function. Distributions with kurtosis less than $3$ are called \textit{platykurtic}, i.e. they produce fewer and less extreme outliers than the normal distribution. Inversely, distributions with kurtosis higher than $3$ are called \textit{leptokurtic} and produce more outliers with respect to the normal distribution. The kurtosis for a class is computed as a weighted average of the kurtosis of the feature values of its samples. The final kurtosis score $\kappa$ represents the average kurtosis over all classes.

\emph{Mutual information}~\cite{kozachenko1987sample,ross2014mutual} measures the mutual dependence between two variables. In the following experiments, it is used to estimate the amount of information obtained about the classification target by observing a data set feature. The overall mutual information score $\eta$ is computed as the average over all features.

\paragraph{Euclidean metrics} Euclidean metrics assess the shape of the data manifold.

The \emph{intrinsic dimensionality ratio} provides a normalized estimate of the dimensionality of the data, considering a linear manifold. It is computed counting the number of principal components needed to explain $90\%$ of the variance in the target~\cite{jolliffe2011principal} ($\mathfrak{I}$). The final score $\mathfrak{I}_r$ is normalized over the number of original features .

\emph{Feature noise} is an estimate of the amount of useless information. Following~\cite{Oreski2017} and~\cite{lopez2013insight}, the score is computed through the difference between dimensionality (the original number of features) and intrinsic dimensionality. The final score $\mathfrak{N}$ is normalized over the original dimensionality.

The \emph{average sample distance} $\mu_\mathfrak{D}$ and the \emph{standard deviation of sample distances} $\sigma_\mathfrak{D}$~\cite{anton2010elementary} measure the average pairwise distance between two data set points, and the standard deviation of the resulting distribution, respectively.

\paragraph{Generalization metrics} Generalization metrics estimate the nature and hardness of the classification task. 
Given the convex hull of a training data set, the ratio of test points inside it $T_{in}$ and outside of it $T_{out}$ assesses the type of generalization task ML models are asked to perform. Indeed, if test points often fall inside the convex hull of the training set, it is expected that the ML model will probably \emph{interpolate} known points to find predicted values, most of the time; vice versa, if test points frequently fall outside the convex hull, the ML model will likely be required to recurrently \emph{extrapolate} to obtain predictions.
Class-imbalance may also play a role in impairing classifier performance. It has been computed for training and test samples, both inside and outiside the convex hull.

\emph{Classification performance} estimates how difficult it is for a given classifier to learn from the training set and generalize to the test set. Since the analyzed data sets usually have more than two classes, the $F1$ score \cite{van1979information} is used to measure the classification performance. More specifically, the weighted $F1$ score is adopted, to account for label imbalance.
Two scores related to the test set are computed, assessing effectiveness in both interpolation ($F1_{in}$) and extrapolation ($F1_{out}$).

\section{Experimental results and discussion}

This section describes the experimental results obtained through the analysis of the selected data sets. First, classical linear correlations between the chosen metrics are considered. Then, more complex non-linear models are explored and discussed, outlining the importance of the convex hull.
While different algorithms might perform differently on the same data set, testing all possible classifying alternatives is impractical. For this purpose, we selected a \emph{Logistic Regression} (LR)~\cite{yu2011dual} classifier, a \emph{Support Vector Machines} classifier with radial basis function kernel (SVC)~\cite{platt1999probabilistic}, and a \emph{Random Forest} classifier with $100$ decision tree estimators (RF)~\cite{breiman2001random} as representative classifiers for the following experiments, taking into account their considerable efficiency and their heterogeneous capacity.

All the code and data necessary to reproduce the experiments is available in a public GitHub repository\footnote{\url{https://github.com/pietrobarbiero/dataset-characteristics}}.

%\begin{itemize}
%    \item analysis of the data sets
%    \item general conclusions
%    \item dimensionality not so cursed
%    \item ficchiamo le nostre ipotesi
%    \item (tipo, tutti i punti sono alla stessa distanza fra loro in alta dimensionalità)
%     \item convex hull, fuori/dentro ha senso solo per dimensioni basse
%     \item forse dovremmo anche mettere qualche metrica, tipo distanza euclidea media fra due punti di tutto il data set
%     \item analisi delle correlazioni che abbiamo trovato
%     \item e anche dell'assenza di correlazione
%     \item discutiamo qualche idea del paper di Google
%     \item IMPORTANTE: mostriamo le correlazioni lineari, le analizziamo, poi diciamo che le correlazioni potrebbero non essere lineari, e proponiamo un altro sistema, Symbolic Regression
% alla luce della nostra analisi empirica
%     \item out-acc non dipende da niente, quindi difficile stimare quanto generalizza bene il modello fuori dal convex-hull del training set; inoltre più è elevata la dimensionalità intrinseca più l'accuratezza sul training è predittiva
%     \item intrinsic-dim-ratio e fcc-mean != 1 $\implies$ train-acc molto buona; se ==1 potrebbe andare molto male (randomforest)
%     \item in-acc non è collegato con out-acc!
% \end{itemize}

\subsection{Correlations between data set characteristics: sometimes dimensionality not so cursed}
% \begin{itemize}
%     \item {\color{red}METTERE FIGURE VECCHIE SU OSF QUI, COME PLACEHOLDERS}
%     \item Proviamo correlazioni classiche dalla letteratura
%     \item Non funzionano (lineari?)
%     \item Proviamo a usare modelli piu' complessi
% \end{itemize}

\begin{figure}[!ht]
    \centering
    \includegraphics[width=.8\textwidth, trim=0cm 0cm 1.25cm 1.25cm, clip=true]{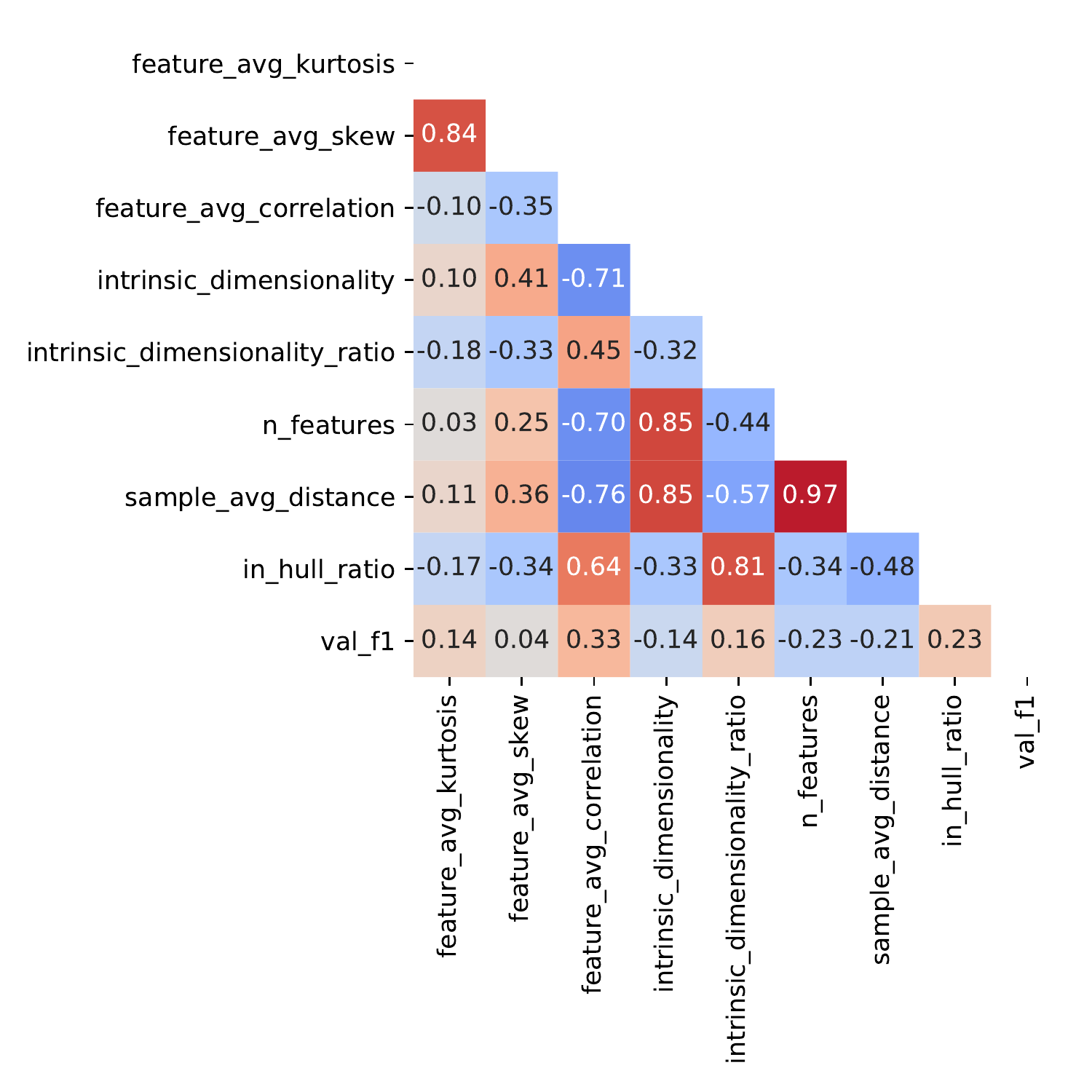}
    \includegraphics[width=0.7\textwidth]{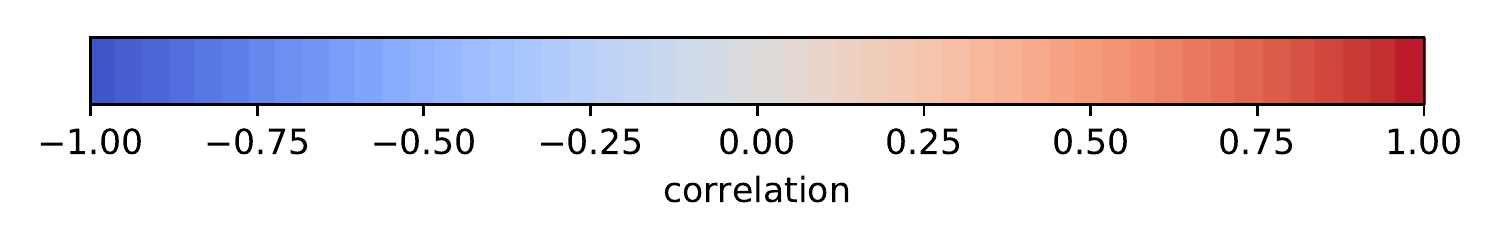}
    \caption{Heatmap showing a selection of the most relevant correlations between data set characteristics. The $F1$ score computed on test samples (\textit{val\_f1}) corresponds to the average score over the three ML models used for the experiments, i.e. LR, SVC, and RF.}
    \label{fig:correlation-matrix-simple}
\end{figure}

Once all the considered metrics described in Sec. \ref{sec:methodology} have been computed for the \numberofdatasets selected data sets, classical statistical correlations can be evaluated. In particular, computing Pearson's correlation coefficients results in the matrix presented in Fig. \ref{fig:correlation-matrix-simple}. Analyzing the matrix, several predictable correlations can be found: In the following, we will analyze a few of the least immediately obvious.

$\gamma\diamond\kappa$: the mean skewness and mean kurtosis of a dataset are highly correlated (0.84), as both metrics assess the difference in data distribution with respect to a reference Gaussian distribution.

$\mathfrak{I}\diamond\rho$: very often, the higher the correlation between features, the lower the intrinsic dimensionality of a dataset, so as expected the two metrics are anti-correlated (-0.71).

$d$: dimensionality is positively correlated with $\mathfrak{I}$ (0.85) and negatively correlated with $\rho$ (-0.70). In other words, as the number of features increase, intrinsic dimensionality tends to rise; and, at the same time, features are less likely to be strongly correlated with each other.

$\mu_{\mathfrak{D}}\diamond\rho$: as the number of features/dimensions increases, the average distance between samples also increases, \emph{unless} the additional features are strongly correlated with existing ones. For this reason, as expected, the average distance between samples is negatively correlated with the average correlation between features (-0.76). For the same reason, we find a 0.85 positive correlation between average sample distance and intrinsic dimensionality ($\mathfrak{I}$), and an equally strong positive correlation between $\mu_{\mathfrak{D}}$ and number of features $d$.

$T_{in}\diamond\mathfrak{I}_r$: the correlation between the ratio of test samples falling inside the convex hull of the training set and the intrinsic dimensionality ratio (0.81) is maybe the less intuitive of the relationships analyzed so far. When the intrinsic dimensionality $\mathfrak{I}$ is lower than the number of dimensions $d$, training points used to compute the convex hull might actually all lie in a polytope of dimension $d'\approx\mathfrak{I} < d$; the likelihood of test points laying exactly inside this polytope becomes then low, especially when compared to a situation where $\mathfrak{I} \approx d$, and the convex hull of the training set occupies a much larger portion of the feature space. The very same concept is also expressed by the negative correlation between $T_{in}$ and $\mathfrak{N}$ (-0.81), as a higher feature noise also represents a lower effective dimensionality. The same reasoning holds for the correlations $T_{out}-\mathfrak{I}_r$ and $T_{out}-\mathfrak{N}$.

Regarding classifier-specific metrics, such as $F1$ for training and validation, we can observe how LR presents the highest correlation between the two, suggesting a similar behavior for training and unseen samples. SVC and RF, on the other hand, have poorer correlations, as they tend to overfit the training set more, as expected by classifiers with higher capacity. It is important to notice that the strength of this correlation does not imply a poor performance, as RF, with the lowest correlation, shows the highest $F1$ on the test set (see Table \ref{tab:res-f1}).

Correlations that are unexpectedly weak in the analysis, those between $F1_{test}$ and all metrics related to dimensionality ($d$, $\mathfrak{I}$, $\mathfrak{I}_r$), hint at a surprising conclusion: The performance of the ML algorithms on an unseen test set is almost independent from the dimensionality of the data set. This is particularly true for LR ($F1_{test}\diamond\mathfrak{I}_r=0.1$) and RF ($F1_{test}\diamond\mathfrak{I}_r=0.9$), while corresponding correlations for SVC are higher, but still not very strong ($F1_{test}\diamond d=-0.43$, $F1_{test}\diamond\mathfrak{I}=-0.37$, but $F1_{test}\diamond \mathfrak{I}_r=0.28$).

For the complete correlation matrices between all considered characteristics, see Figs.~\ref{fig:correlation-matrix-lr}, \ref{fig:correlation-matrix-svc}, \ref{fig:correlation-matrix-rf}, in Appendix \ref{appendix:correlation-matrices}.

\subsection{The key role of the convex hull and association models}

In Table \ref{tab:res-f1} we reported $F1$-scores of ML models on training and validation sets. A high difference between $F1_{train}$ and $F1_{test}$ corresponds to overfitting.
We record RF exhibiting higher overfitting, while still providing the best generalization performance during validation.
On the other hand, the difference between $|F1_{train}-F1_{in}|$ and $|F1_{train}-F1_{out}|$ reveals the discrepancy between interpolation and extrapolation performances, pinpointing the importance of the convex hull in assessing machine learning generalization.

% Please add the following required packages to your document preamble:
% \usepackage{multirow}
% \usepackage{graphicx}
\begin{table}[!ht]
\renewcommand{\arraystretch}{1.5}
\centering
\caption{Average $F1$-score and standard error of the mean of ML models.}
\label{tab:res-f1}
% \resizebox{\textwidth}{!}{%
\begin{tabular}{llll}
\hline
ML model & Sample set & $F1$-score & $\Delta$ \\ \hline
\multirow{4}{*}{LR} & $F1_{train}$ & $0.87\pm 0.01$ &  \\
 & $F1_{test}$ & $0.79\pm 0.01$ & $0.08 \pm 0.03$ \\
 & $F1_{in}$ & $0.82\pm 0.01$ & $0.05 \pm 0.03$ \\
 & $F1_{out}$ & $0.78\pm 0.01$ & $0.09 \pm 0.03$ \\
 \hline
\multirow{4}{*}{SVC} & $F1_{train}$ & $0.86 \pm 0.00$ &  \\
 & $F1_{test}$ & $0.78 \pm 0.01$ & $0.08 \pm 0.01$ \\
 & $F1_{in}$ & $0.85 \pm 0.01$ & $0.01 \pm 0.01$ \\
 & $F1_{out}$ & $0.76 \pm 0.01$ & $0.10 \pm 0.01$ \\
 \hline
\multirow{4}{*}{RF} & $F1_{train}$ & $0.99\pm 0.00$ &  \\
 & $F1_{test}$ & $0.84\pm 0.00$ & $0.15 \pm 0.01$ \\
 & $F1_{in}$ & $0.88\pm 0.00$ & $0.11 \pm 0.01$ \\
 & $F1_{out}$ & $0.82\pm 0.00$ & $0.17 \pm 0.01$\\
\hline
\end{tabular}%
% }
\end{table}

\begin{figure}[!ht]
    \centering
    \includegraphics[width=0.49\textwidth]{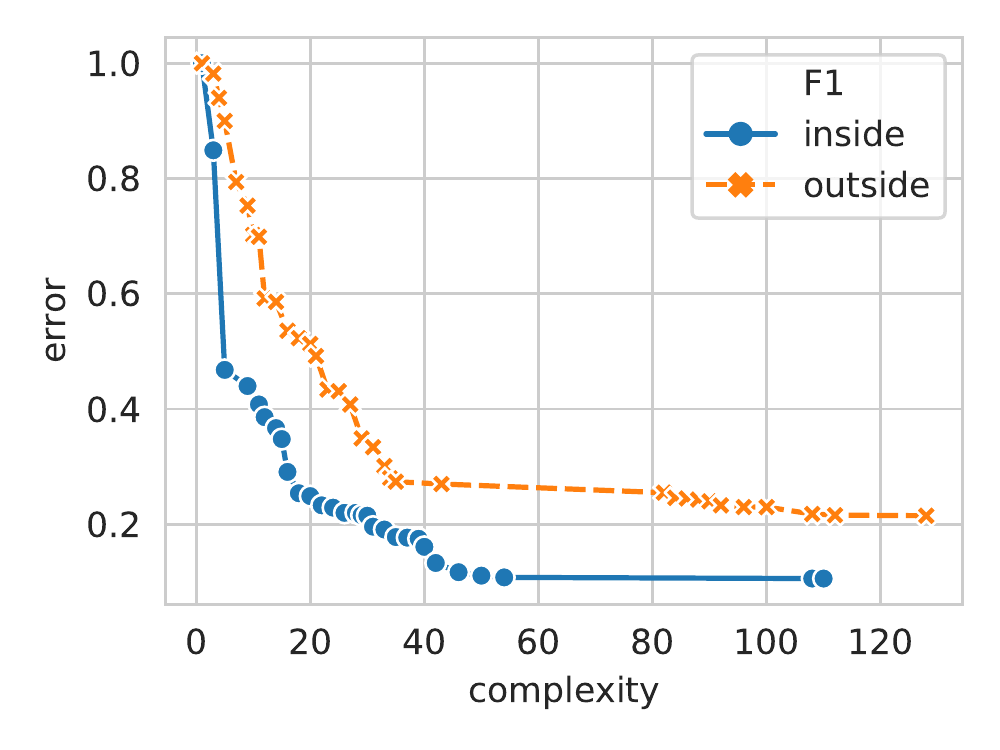}
    \includegraphics[width=0.49\textwidth]{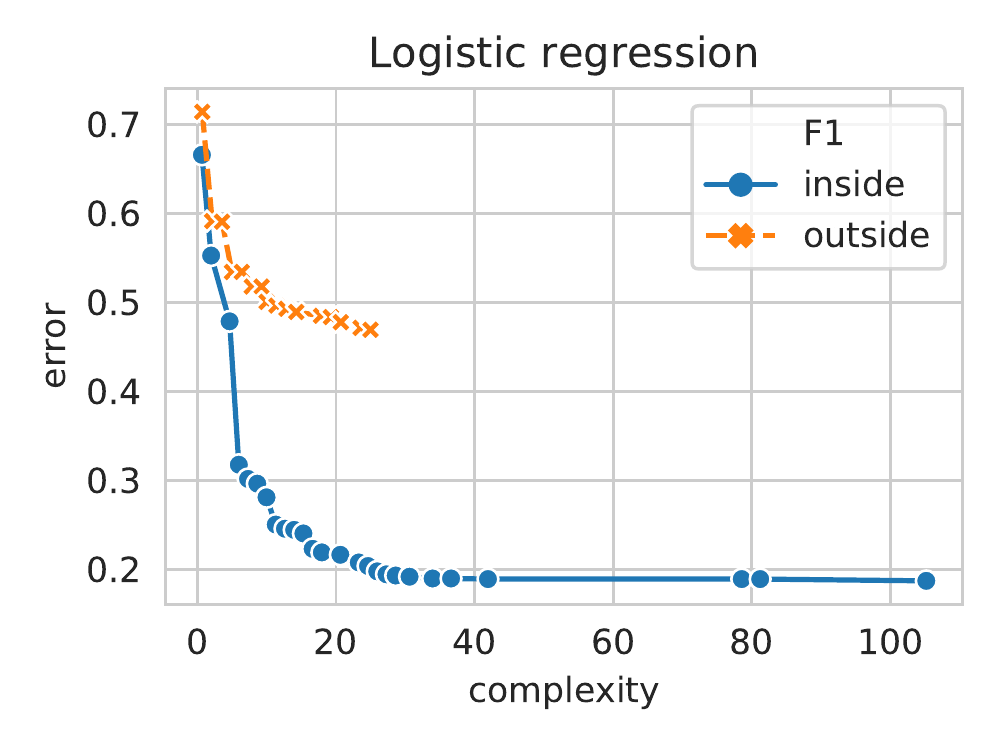}\\
    \includegraphics[width=0.49\textwidth]{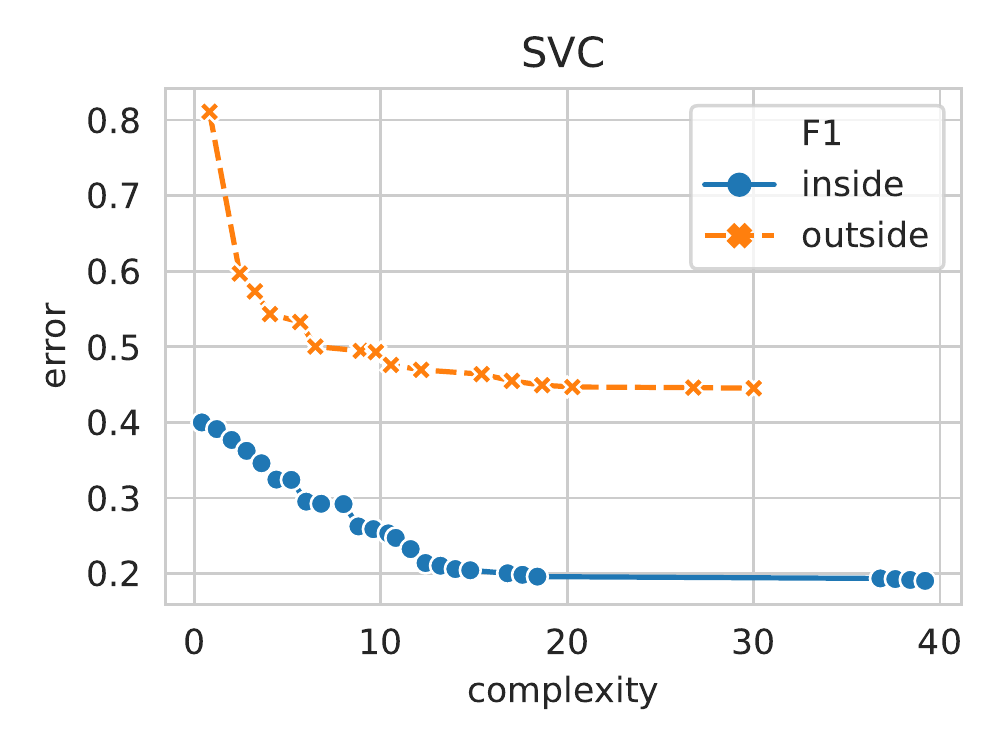}
    \includegraphics[width=0.49\textwidth]{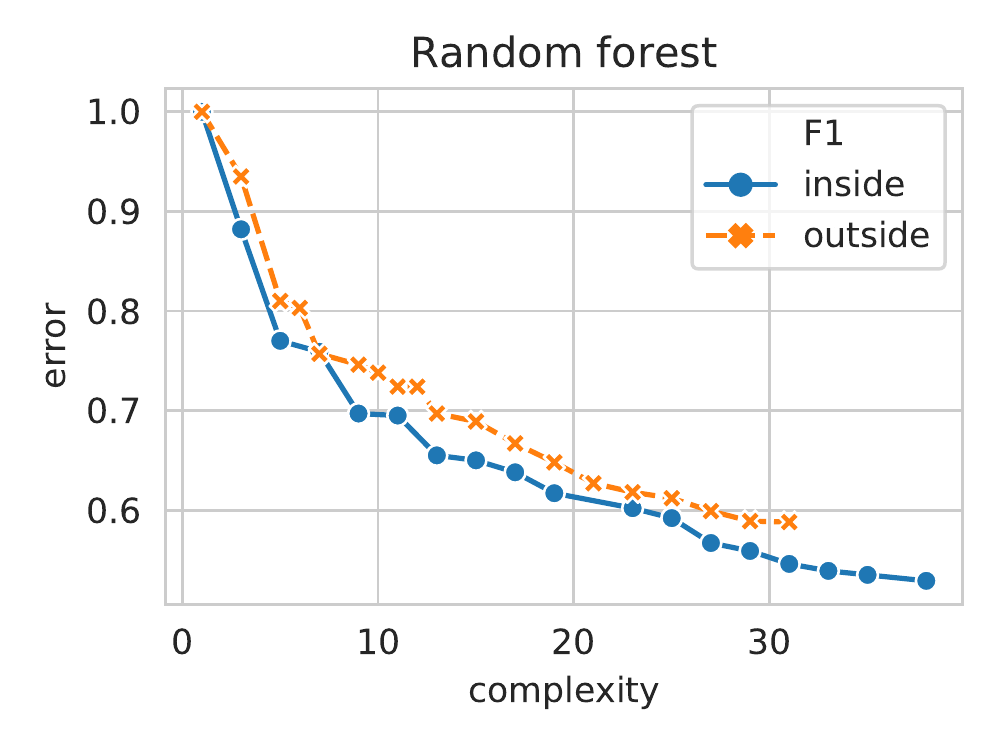}
    \caption{Pareto-optimal models predicting $F1_{in}$ and $F1_{out}$ based on data set characteristics taking into account the results on all ML models (\textbf{top-left}). Pareto fronts for each ML model: logistic regression (\textbf{top-right}), SVC (\textbf{bottom-left}), and random forest (\textbf{bottom-right}).}
    \label{fig:res1}
\end{figure}

\begin{figure}[!ht]
    \centering
    \includegraphics[width=0.49\textwidth]{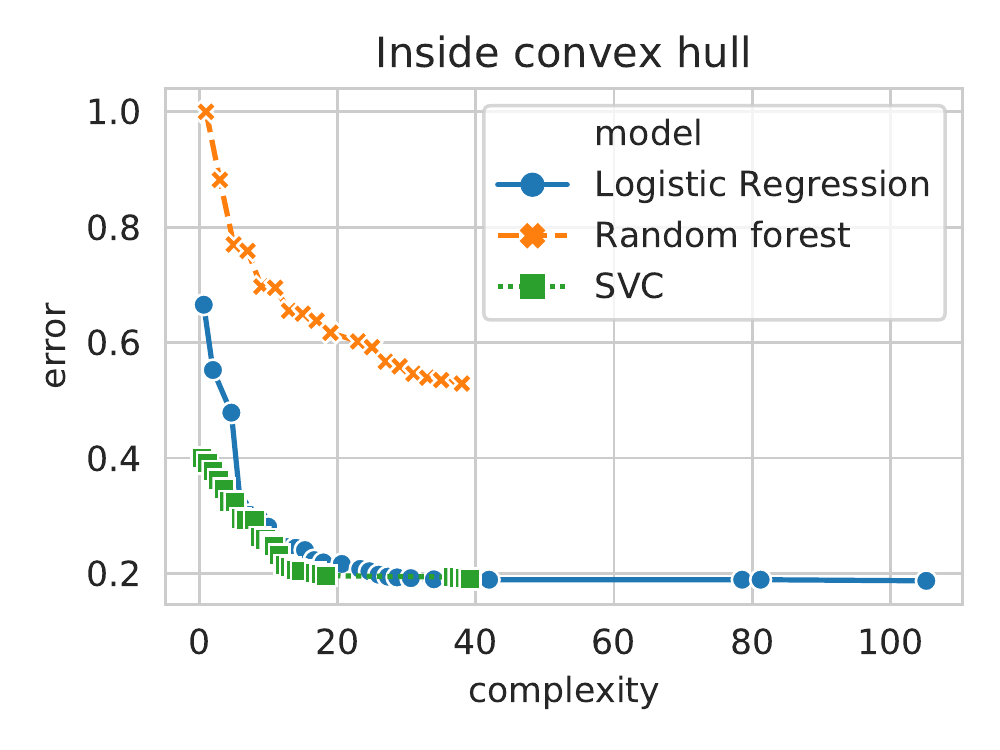}
    \includegraphics[width=0.49\textwidth]{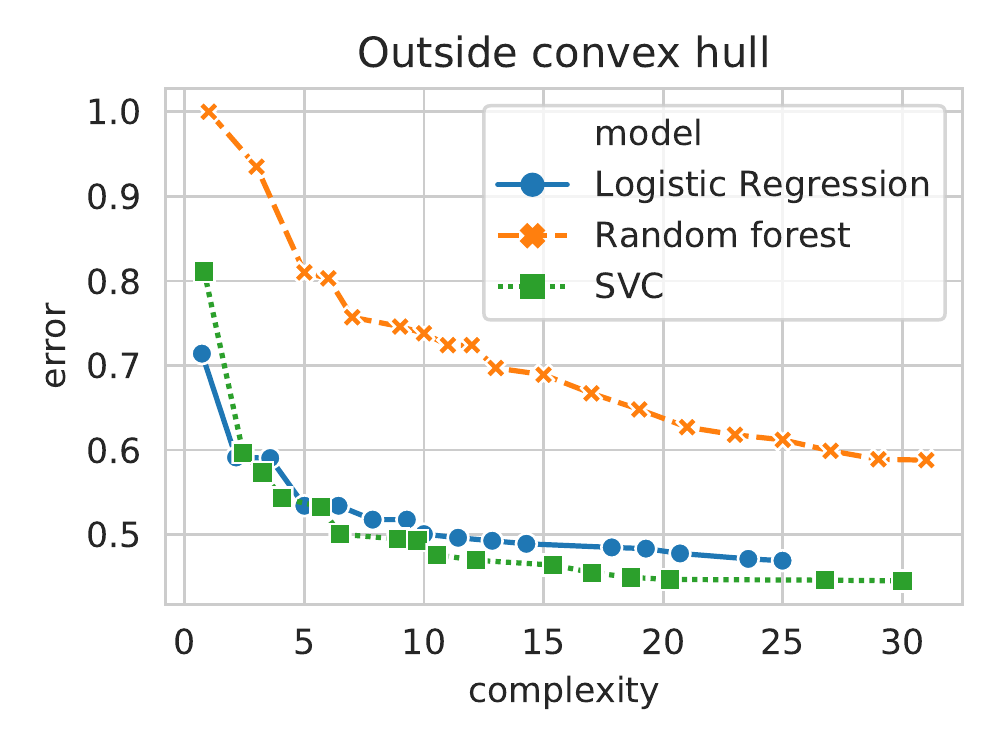}
    \caption{Comparison of Pareto fronts predicting $F1_{in}$ and $F1_{out}$ based on data set characteristics.}
    \label{fig:res2}
\end{figure}

While classical correlation metrics try to optimize the coefficients of models of known structure (e.g., often linear), it might be useful to extend such analysis to models of different structure. In statistical terms, this implies assessing \emph{association}, a relationship between variables more general than correlation: Two or more variables are \emph{associated} if the values of some provide information on the value of the others~\cite{Altman2015}. To this purpose, we propose the use of \emph{symbolic regression}~\cite{koza1992genetic,schmidt2009distilling}, a technique that searches the space of mathematical expressions to find the model that best fits a given data set. The irony of using machine learning techniques to analyze the results of a meta-analysis on machine learning is not lost on the authors, but symbolic regression has the advantage of returning completely human-readable models, that can later be interpreted and explained, all the while considering relationships between data set characteristic more complex than just linear correlations. Furthermore, symbolic regression can deliver multiple candidate solutions, models of increasing complexity and fitting, whose meta-analysis can deliver additional information to the user. For this task, the commercial symbolic regression software Eureqa Formulize\footnote{Eureqa Formulize is developed by Nutonian, Inc. \url{https://www.nutonian.com/products/eureqa/}} is employed. All available building blocks for equations were selected, with the exception of those specifically designed for time series analysis. Each run is stopped when the convergence metric of Eureqa crosses the threshold of 90\%.%, we analyzed and compared the Pareto front obtained through the application of symbolic regression to the characteristics extracted from the \numberofdatasets data sets analyzed. 

In order to assess generalization abilities of machine learning models, we analyze and compare the results provided by Eureqa in different scenarios. %For each classifier (i.e. RF, SVC, and LR) we use symbolic regression in order to explore the associations between data set characteristics and classification performances, for samples inside and outside the convex hull of the training set. % 
For each classifier, symbolic regression generates a set of Pareto-optimal models, predicting the performance of the classifier in terms of $F1$-score. Pareto optimality is considered as a function of both $R^2$ (i.e. the accuracy of the formula) and complexity (the number of terms and complexity of formula's building blocks).
Among all Pareto-optimal equations proposed by symbolic regression, we manually analyze a selection representing reasonable compromises between fitting and complexity. Eqs. \ref{eq:f1-in-lr}-\ref{eq:tin} represent candidate equations of similar complexity ($C=9$), describing nonlinear associations between data set characteristics and generalization metrics for each machine learning classifier.

\begin{equation} \label{eq:f1-in-lr}
    F1_{in}^{LR} = 0.857 + (0.959 \cdot F1_{train}^{LR} - 0.831) \ \text{step}(2.38 \cdot \mathfrak{I}_r - 0.0019) \qquad [R^2=0.68,C=9]
\end{equation}

\begin{equation}
    F1_{out}^{LR} = 0.843 \cdot F1_{train}^{LR} + 0.176 \cdot \lambda - 0.040 - 0.097 \cdot \mathfrak{N} \qquad [R^2=0.47,C=9]
\end{equation}

\begin{equation}
    F1_{in}^{SVC} = 1.072 \cdot F1_{train}^{SVC} + 0.021 \cdot \sigma_{\mathfrak{D}} - 0.08 - 0.003 \cdot c \qquad [R^2=0.67,C=9]
\end{equation}

\begin{equation}
    F1_{out}^{SVC} = 0.820 + (1.605 \cdot F1_{train}^{SVC} - 1.389) \ \cos(5.424 + 2.377 \cdot \mathfrak{I}_r) \qquad [R^2=0.45,C=9]
\end{equation}

\begin{equation}
    F1_{in}^{RF} = 7.66 \cdot \rho - 1.822 - 5.266 \cdot \rho^2 \qquad [R^2=0.36,C=9]
\end{equation}

\begin{equation}
    F1_{out}^{RF} = 0.818 + (1.237 \cdot \rho - 0.767) \ \text{erfc}(4.65 \cdot \rho - 2.883) \qquad [R^2=0.28,C=9]
\end{equation}

\begin{equation} \label{eq:tin}
    T_{in} = 0.46 + 0.42 \cdot \text{erf}(4.65 \cdot \rho - 2.20 - 2.98 \cdot \mathfrak{N}) \qquad [R^2=0.88,C=9]
\end{equation}

Overall, the interpolation ability ($F1_{in}$) of a ML algorithm on a data set can be predicted in a satisfying way using only the data set characteristics that we analyzed in this study. On the contrary, predicting extrapolation ability ($F1_{out}$) from data set characteristics seems much harder, as pointed out by the lower $R^2$ scores of models for corresponding complexity. Besides, the difference between predictors found for LR or SVC, with respect to RF, is noteworthy: In fact, the generalization ability of the models for the first two algorithms seems strongly associated with their training performance ($F1_{train}$) and either the intrinsic dimensionality ratio ($\mathfrak{I}_r$) or the feature noise ($\mathfrak{N} = 1 - \mathfrak{I}_r$). On the other hand, the test accuracy of RF seems much harder to predict, and associated with the average class-wise correlation among features ($\rho$) \textit{only}.
Moreover, we observe how the ratio of test samples falling inside the convex hull could be easily estimated by considering the class-wise feature correlation ($\rho$) and either the feature noise ($\mathfrak{N}$) or the intrinsic dimensionality ratio ($\mathfrak{I}_r = 1 - \mathfrak{N}$), confirming the reasoning derived from the previous analysis of the linear correlations: Higher feature noise leads to a lower effective dimensionality, thus reducing the likelihood of test samples falling inside the convex hull.

It is interesting to remark how $F1_{train}$ is the variable that explains most of the variance for LR and SVC models; while the same variable does not appear in models for RF, showing how the generalization ability of this classifier is poorly correlated with its performance on the training set, it is generally harder to predict (lower $R^2$), and seems to solely depend on the class-wise feature correlation $\rho$. The model for $T_{in}$ also presents interesting insights, displaying rather high $R^2=0.88$ and depending on just two variables, $\rho$ and $\mathfrak{N}$. The negative influence of $\mathfrak{N}$ can be explained intuitively: As the feature noise increases, the intrinsic dimensionality ratio reduces, and thus the points belonging to the convex hull of the training set lie more and more on a polytope of lower dimensionality than the entire feature space, making it more difficult for test points to fall inside its hypervolume. The positive influence of $\rho$ on $T_{in}$ is harder to explain: We speculate that a higher class-wise feature correlation would bring points belonging to the same class closer together in the feature space, as depicted in Fig.~\ref{fig:data-sparsity}. Having the data points gathered in a smaller part of the feature space might imply that more test points will fall inside the convex hull of the training set, without necessarily reducing the true dimensionality of the feature space. This is somehow confirmed by the results reported in Fig.~\ref{fig:correlation-matrix-simple}, where intrinsic dimensionality ratio and average class-wise feature correlation are positively correlated, albeit weakly ($\mathfrak{I}_r\diamond\rho=0.45$).

% TODO add them
% F1out: i modelli diventano molto complicati per avere una R2 decente (vedi figura XXX), il ché significa che non esistono associazioni semplici fra le caratteristiche considerate, capaci di spiegare F1out; inoltre, anche con modelli molto complicati che tengono in considerazione $\mathfrak{I}_r$, l'R2 è comunque 0.69, che non è niente di eccezionale.
% E' possibile rifare l'esperimento usando solo i dati di LR e RF?
% An automatically generated meta-analysis of the models found by Eureqa is reported in Appendix...

We further extend the analysis of symbolic regression results by comparing Pareto fronts of ML models both inside and outside the convex hull to further inspect whether data set characteristics have a significant impact on models' performances. If Pareto front \textit{A} dominates Pareto front \textit{B} it is likely that data set characteristics have a higher impact on ML performances in the first scenario rather than in the second one.
The Pareto front analysis is presented in Figs.~\ref{fig:res1} and \ref{fig:res2}. In Fig.~\ref{fig:res1}, we analyze the link between data set characteristics and generalization ability by comparing \textit{interpolation} and \textit{extrapolation} results of ML models. In all scenarios the relationship looks stronger when predictions are made inside the convex hull of training samples. However, while this difference is emphasized for LR and SVC, it is less pronounced for RF. 
In Fig.~\ref{fig:res2} we further inspect symbolic regression results by comparing Pareto fronts inside and outside the convex hull. Once more, we observe stronger associations between data set characteristics and LR or SVC compared to RF. This means that the impact of data set-specific properties on RF performances is lower, as if the higher capacity of the model makes it more robust.

\bigskip
\begin{tcolorbox}[colback=blue!5!white,colframe=blue!75!black,title=Highlights of experimental findings]
\begin{itemize}
    \item The convex hull has a key role in assessing ML algorithm generalization, in terms of interpolation and extrapolation abilities.
    \item The structure and fitting of the association models found for $F1_{in}$ suggests that predicting the interpolation ability of a ML algorithm on a data set might be feasible, using only the data set characteristics that we analyzed in this study. On the contrary, after analyzing the association models obtained for $F1_{out}$, predicting extrapolation abilities from data set characteristics seems much harder.
    \item ML models with high capacity seem to generalize better (both inside and outside the convex hull), despite their higher potential for overfitting. The phenomenon is similar to what has already been observed in neural networks by Zhang et al. \cite{zhang2016understanding}.
    \item The performance of the ML algorithms on an unseen test set seems almost independent from the dimensionality of the data set, thus challenging the common assumption that the \textit{curse of dimensionality} might impair generalization in machine learning.
\end{itemize}
\end{tcolorbox}
\bigskip

\section{Conclusions}

After presenting our analysis of the correlations found on the \numberofdatasets data sets we analyzed, there is a (rather ironic) question we have to face: how \emph{general} are the results we found? Or, in other words, how well do the \emph{predictions} we perform \emph{extrapolate} to unknown data sets? Frankly speaking, we cannot claim that the correlations described in this work hold for all possible data sets, but the sheer number of different data sets analyzed gives us some hope of generality.

There is, however, a possible bias in the selection of data sets for this study: we focused on openly accessible, curated data sets, that already had to pass several quality checks in order to be hosted on repositories such as OpenML. This pre-selection process might make the data sets considered in this work not representative of all real-world data sets. In other words, usually a data set is uploaded on OpenML because the authors already know that at least one ML technique is going to work well for that specific data set; thus, what we analyzed might be representative only of data sets for which ML techniques work well.

Another possible explanation for the most counter-intuitive correlations we uncovered is that real-world data sets are a subset of all possible data sets. While some general mathematical conclusions, such as the curse of dimensionality, might hold for the set of all hypothetical data sets, they might not necessarily be true for the subset of data that is measured from real phenomena. This observation mirrors the remarks by Lin et al. \cite{lin2017does}: In an attempt to explain the effectiveness of neural networks and ML at representing physical phenomena, the authors notice that laws of physics can typically be approximated by a tiny subset of all possible polynomials, of order ranging from 2 to 4; this is a consequence of such phenomena usually being symmetrical when it comes to rotation and translation. As the data sets we analyzed come from either simulations or real-world experiments, their characteristics might lead ML algorithms to represent them more easily than expected.

\bibliographystyle{unsrt}  
\bibliography{sample,addenda}  %%% Remove comment to use the external .bib file (using bibtex).

\clearpage

%% The Appendices part is started with the command \appendix;
%% appendix sections are then done as normal sections
\appendix

\section{Code}
\label{appI}

\begin{lstlisting}[language=Python, caption=Python implementation of the convex hull test]
import numpy as np
from scipy.optimize import linprog

def convex_hull_check(X_train, x_test):
    n_points = len(X_train)
    c = np.zeros(n_points)
    A = np.r_[X_train.T, np.ones((1, n_points))]
    b = np.r_[x_test, np.ones(1)]
    lp = linprog(c, A_eq=A, b_eq=b)
    return lp.success
\end{lstlisting}

\clearpage

\section{Full correlation matrices}
\label{appendix:correlation-matrices}
\begin{figure}[!ht]
    \centering
    \includegraphics[scale=0.3]{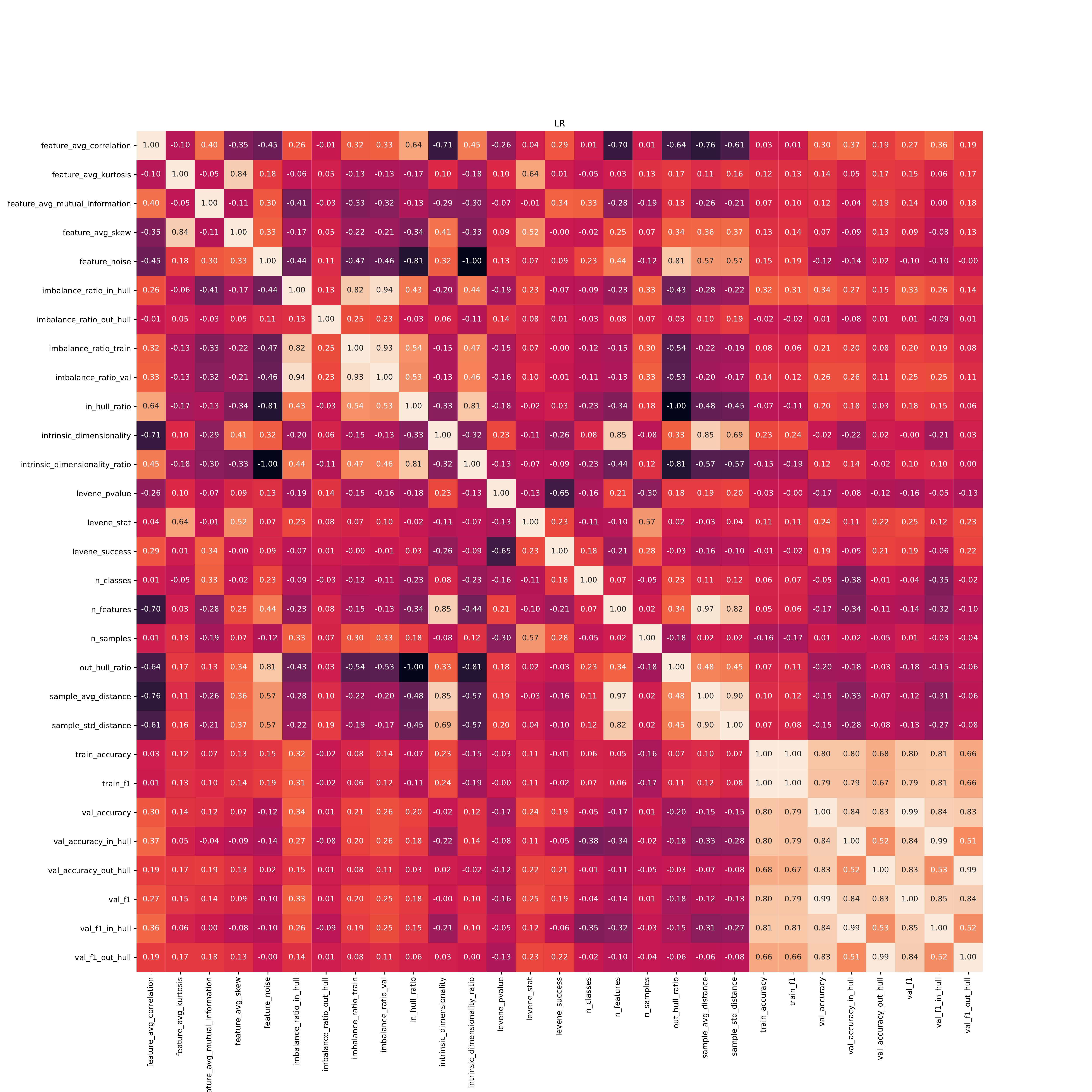}
    \caption{Correlations between dataset characteristics using logistic regression.}
    \label{fig:correlation-matrix-lr}
\end{figure}

\begin{figure}
    \centering
    \includegraphics[scale=0.3]{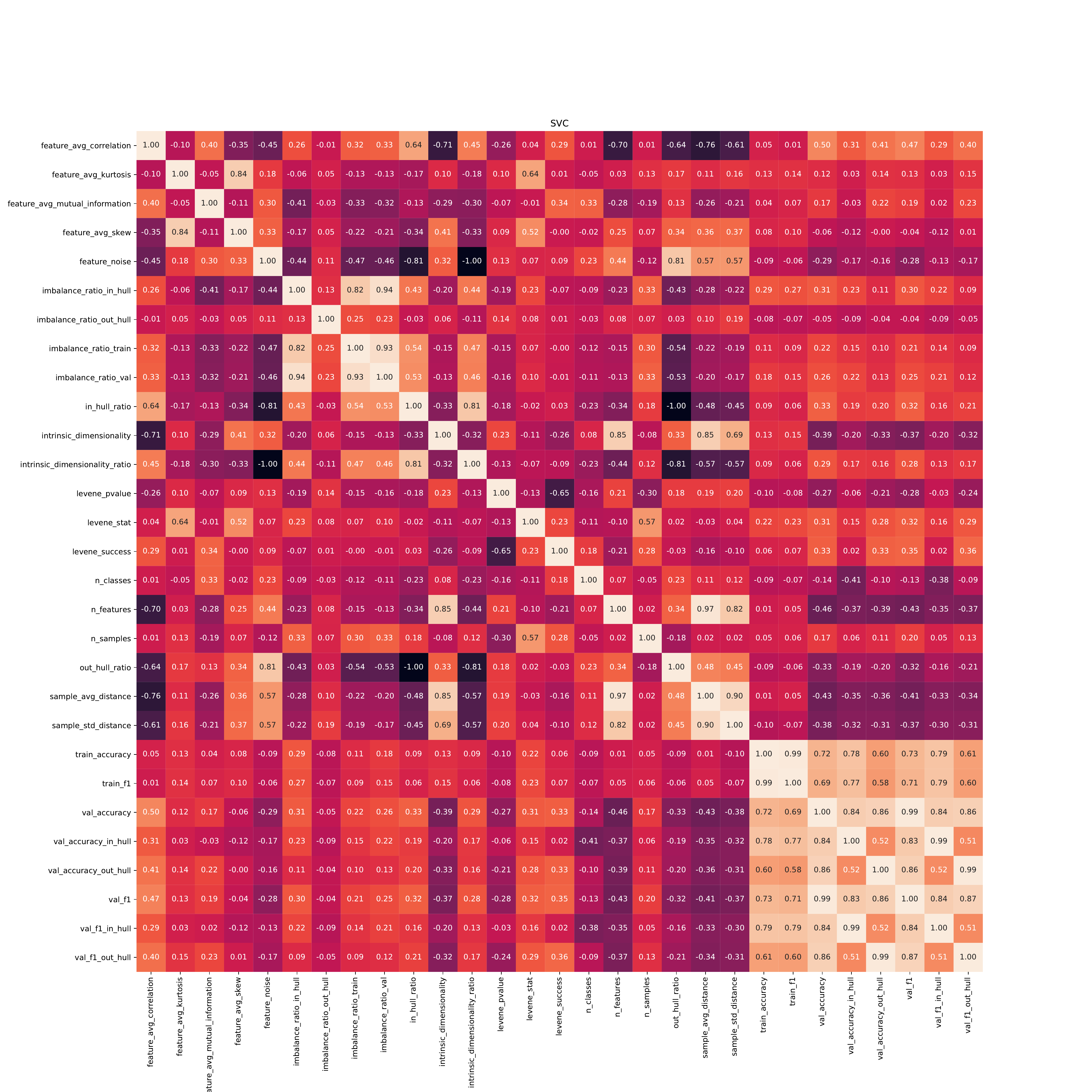}
    \caption{Correlations between dataset characteristics using SVC.}
    \label{fig:correlation-matrix-svc}
\end{figure}

\begin{figure}
    \centering
    \includegraphics[scale=0.3]{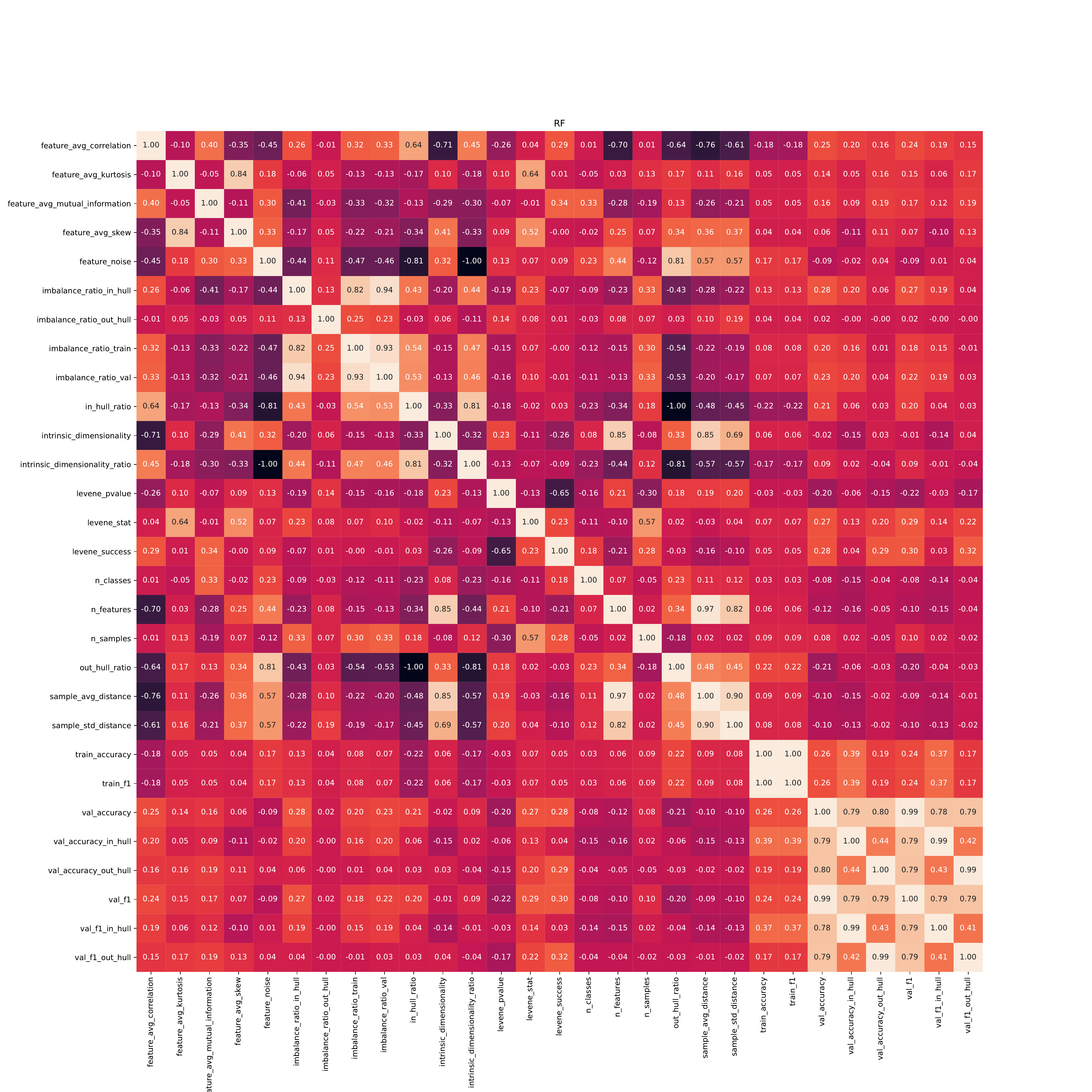}
    \caption{Correlations between dataset characteristics using random forest.}
    \label{fig:correlation-matrix-rf}
\end{figure}

\end{document}